\documentclass[11pt,a4paper]{article}

\usepackage[T1]{fontenc}
\usepackage[utf8]{inputenc}
\usepackage[margin=1in]{geometry}
\usepackage{graphicx}
\usepackage{amsmath}
\usepackage{amssymb}
\usepackage{array}
\usepackage{xcolor}
\usepackage[colorinlistoftodos]{todonotes}
\usepackage{authblk}
\usepackage{fancyhdr}
\usepackage[colorlinks=true,linkcolor=blue,citecolor=blue,urlcolor=blue]{hyperref}

\newcommand{\lcnote}[1]{}
\newcommand{\ennote}[1]{}
\definecolor{pdcolor}{rgb}{0.00,0.45,0.20}
\newcommand{\pdnote}[1]{}
\definecolor{fixcolor}{rgb}{0.87,0.45,0.00}

\AtBeginDocument{%
  }

\newcommand{\keywords}[1]{\par\smallskip\noindent\textbf{Keywords: }#1\par\smallskip}

\begin{document}

\title{LAAF: A Layered Accountability Architecture Framework for LLM Applications}

\author[1]{Prachi Chaturvedi\thanks{\texttt{prachic1811@gmail.com}, ORCID~0009-0003-8411-6050}}
\author[1]{Shahnawaz Ahmad\thanks{\texttt{shahnawaz98976@gmail.com}, ORCID~0000-0002-3989-9928}}
\author[2]{Ehsan Nowroozi\thanks{\texttt{ehsan.nowroozi65@ieee.org}, ORCID~0000-0002-5714-8378}}
\author[2]{Muhammad Waqas\thanks{\texttt{muhammad.waqas@gre.ac.uk}, ORCID~0000-0003-0814-7544}}
\author[2]{George Loukas\thanks{\texttt{g.loukas@gre.ac.uk}, ORCID~0000-0003-3559-5182}}
\author[3]{Alireza Jolfaei\thanks{\texttt{alireza.jolfaei@curtin.edu.au}, ORCID~0000-0001-7818-459X}}
\author[4]{Lucas Cordeiro\thanks{\texttt{lucas.cordeiro@manchester.ac.uk}, ORCID~0000-0002-6235-4272}}
\author[4]{Pierre Dantas\thanks{\texttt{pierre.dantas@manchester.ac.uk}, ORCID~0000-0001-6390-9340}}

\affil[1]{School of Computer Science Engineering and Technology, Bennett University, Greater Noida, India}
\affil[2]{Centre for Sustainable Cyber Security (CS2), University of Greenwich, United Kingdom}
\affil[3]{School of Electrical Engineering, Computing \& Mathematical Sciences, Curtin University, Australia}
\affil[4]{Systems and Software Security (S3), University of Manchester, United Kingdom}

\date{}

\maketitle

\begin{abstract}

Large Language Models (LLMs) operate in hospitals, courtrooms, banks, and public-service desks, where fluent, confident outputs are treated as authoritative even when ungrounded or incorrect. When such an output contributes to harm, who is answerable, and through what mechanisms can responsibility be traced, explained, and acted upon? This review surveys how the literature answers that question and maps it against the instruments now in force. Following PRISMA reporting guidance, five databases were searched for the period January 2022 to March 2026 against four review questions; of 4,512 records identified, 122 primary studies were included, together with 12 regulatory and standards documents analysed as primary sources. The review consolidates a sociotechnical account of accountability as an actor--forum relation resolved into five dimensions, and synthesises mechanisms across four families spanning technical controls, human oversight, organisational governance, and documentation and traceability, each with a maturity assessment. The corpus is read throughout through a four-layer classification device spanning provenance, application logic, human oversight, and governance and redress, cross-cut by traceability, role clarity, and continuous monitoring. Both are mapped onto the EU AI Act, whose high-risk obligations have applied since 2 August 2026, the NIST AI RMF with its Generative AI Profile, ISO/IEC 42001, and sectoral guidance in healthcare, consumer finance, education, and the public sector. Four persistent gaps emerge, namely under-specification of human oversight, absence of shared accountability metrics, disciplinary disconnection, and limited empirical evaluation, alongside five structural tensions that no surveyed instrument resolves. The review closes by consolidating the classification device into an integrated accountability architecture, LAAF, with cybersecurity aligned to the OWASP LLM Top 10 (2025); it is a synthesis of the surveyed evidence rather than a validated artefact. To our knowledge this is the first systematic review to map the LLM accountability literature onto the binding regulatory instruments now in force.

\end{abstract}


\section{Introduction}
\subsection{The Accountability Imperative for LLMs}

A few years ago, asking a computer a question meant typing keywords into a search box and waiting for a list of links. Today, a doctor can type a clinical question into a chatbot and receive a paragraph-long answer that reads as if written by a senior colleague \cite{li2024chatgpt,he2025survey}.The banker can request a credit memo, the teacher a lesson plan, and the public servant a policy note, and in each case the response arrives in fluent, well-formed prose \cite{jiao2025generative}. Assistants do hedge and decline, so the problem is not uniform assurance but the absence of any reliable tie between the register of a response and whether it is grounded. Fluency of this kind invites reliance whether or not reliance is warranted, and reliance raises the question this survey addresses: when such an output contributes to harm, who is answerable, and through what mechanisms?

The difficulty begins with a structural asymmetry. A database that holds no matching record returns nothing; an LLM always returns something, including when it lacks the knowledge the question requires \cite{huang2025survey}. Without adequate grounding it produces hallucinations, that is, statements that are delivered in the same register as supported ones, syntactically well formed, and factually wrong \cite{huang2025survey,wang2025survey}. In casual use a hallucination is a nuisance; in high-stakes settings it is a harm. Instances of particular systems failing are plentiful but are the weakest
available evidence, since hallucination rates fall across releases and an
undated instance invites the reply that the defect has been engineered away.
Two formal results bound the problem instead. \cite{kalai2024calibrated} show
that a model calibrated to its training distribution must hallucinate at a rate
bounded below by the fraction of facts appearing once in that distribution, so
calibration and factuality are in tension by construction.
\cite{xu2024hallucination} argue from computability that any computable LLM
admits inputs on which it must hallucinate, irrespective of architecture, data,
or scale. Neither implies mitigation is futile; jointly they establish that
hallucination is a property of the generative regime rather than a transient
defect, and must therefore be governed rather than waited out. Empirical
evaluations then serve a narrower purpose, measuring what the failure costs
where it lands: fabricated references in evidence synthesis \cite{chelli2024hallucination}, invented precedent in legal work \cite{dahl2024large}, and biased recommendations in hiring and lending that shape opportunities unnoticed \cite{abbas2025lending, gallegos2024bias}.

Unlike conventional software defects, LLM errors are emergent properties of probabilistic systems trained on large, heterogeneous corpora, and they cannot be located and patched in the way a bug can \cite{huang2025survey,zhao2026survey}. A system that passes inspection on one day may behave dangerously the next, following a model update, a change to the retrieval corpus, or a shift in deployment context \cite{leon2026lifecycle}. Capability has therefore advanced faster than the mechanisms that would make its use answerable: deployments routinely lack a named accountable actor, a traceable record of how a given output was produced, and a defined route to correction and redress.

The vocabulary of responsible Artificial Intelligence (AI) has grown to include transparency, fairness, explainability, robustness, and trustworthiness, but none of these terms alone answers the simple governance question: if this system causes harm, who is responsible and what happens next? \cite{hollanek2023ai,novelli2024accountability,ferdaus2026towards}. That is a question of accountability, and although its components have each been surveyed separately, no survey has yet brought them into a single account of who answers for what, which is the gap this review fills.

\subsection{The Accountability Gap in Current Literature}
\label{sec:gap} %label added by PD

The literature exhibits four structural characteristics that this review investigates systematically. They are named here to orient the reader and are re-established against the corpus in Section 8.5, where each is reported as a finding of the review; Table~\ref{tab:related_work} substantiates the framing by scoring fourteen recent surveys and major contributions against the dimensions this review integrates.

Four characteristics recur. Human oversight is under-specified, invoked without stating which human reviews, on what evidence, and with what authority to overrule. Shared accountability metrics are absent, the technical literature measuring hallucination and bias while the governance literature measures neither oversight nor redress. The technical and governance literatures are disciplinarily disconnected and neither addresses sectoral variation. And empirical evaluation is limited, most mechanisms being argued for rather than measured \cite{laux2024institutionalised, novelli2024accountability, sharma2026human, raji2020closing,yehudai2025survey}.

Underlying all four is a conceptual conflation: accountability is treated as interchangeable with transparency, explainability, or liability, when it is none of them. Transparency without accountability is publication without consequence, explainability without accountability is description without judgement, and liability without accountability is punishment without proof \cite{hollanek2023ai,novelli2024accountability}. Section 5.1 develops this distinction and uses it as the basis for the working definition adopted throughout the paper.

\subsection{Review Questions}

The four characteristics of Section 1.2 translate into four review questions (RQs). The conceptual conflation motivates RQ1, the absence of shared metrics and the under-specification of oversight motivate RQ2, the unmapped regulatory landscape motivates RQ3, and the disciplinary disconnection motivates RQ4. 

\textbf{RQ1:} How is accountability defined and operationalised in the existing literature on LLM-based applications, and what conceptual gaps remain? 

\textbf{RQ2:} What technical, human-oversight, organisational, and documentation mechanisms support accountability in LLM applications, and how mature is each?

\textbf{RQ3:} How do these mechanisms map onto the emerging regulatory and standards landscape, particularly the EU AI Act, the NIST AI RMF, and ISO/IEC 42001, and where are the gaps? 

\textbf{RQ4:} What patterns and gaps persist across the mechanisms, regulatory
requirements, and sectoral practices surveyed, and how can they be organised
into an integrated account of accountability across the LLM application lifecycle? 

\subsection{Related Surveys and Prior Work}

Recent surveys in this venue address adjacent but distinct concerns: security and privacy at the level of attack surfaces and defences \cite{das2025security}, factuality at the level of output properties \cite{wang2025survey}, shared-training-data detection \cite{naser2025auditing}, and ethical or robust LLMs at the level of principles \cite{ferdaus2026towards}; none maps these onto binding obligations or onto the actor who answers when a failure causes harm, which is the gap this review addresses.

Outside this venue, the clearest available conceptual account of AI accountability \cite{novelli2024accountability} stops at the concept, offering neither a synthesis of the mechanisms that would realise it nor an allocation of it to actors. A three-layered audit procedure for LLMs \cite{mokander2024auditing} addresses auditing rather than the full accountability relation, is not mapped onto the enacted regulatory instruments, and treats neither cybersecurity nor sectoral calibration. Risk governance supplies the organisational apparatus through which answerability is institutionalised \cite{schuett2025three} without connecting it to the technical layer, and the attributability gap is identified \cite{zeiser2024owning} without extension to the mechanisms that would close it.

This review builds on all of these, and its angle is integrative rather than incremental. To our knowledge it is the first systematic review to read the accountability literature through a single layered classification device, to map that literature onto the EU AI Act, the NIST AI RMF, and ISO/IEC 42001 together, and to treat conceptual foundations, mechanism synthesis, regulatory mapping, sectoral calibration, and cybersecurity within one corpus. Table~\ref{tab:related_work} records the comparison; the scoring rubric is given in Section 3.7.

\subsection{Contributions of This Review}
\label{sec:contributions}

This review makes five contributions.

\begin{itemize}
    \item A consolidated conceptual treatment of accountability for LLM-based applications, drawing on AI governance, applied ethics, computer science, and regulatory studies, and resolving accountability into five analytically distinct dimensions with their associated failure modes (Section 5, Table~\ref{tab:five_dimensions}).

 \item A systematic synthesis of accountability mechanisms across four families, namely technical controls, human oversight, organisational governance, and documentation and traceability, with a maturity assessment and regulatory anchor for each family (Section 6, Table~\ref{tab:mechanism_families}).

\item A comparative regulatory mapping of the literature onto the EU AI Act, the NIST AI RMF with its GenAI Profile, ISO/IEC 42001, and sectoral guidance in four sectors: healthcare (FDA, EMA), consumer finance (CFPB, EBA), education (US Department of Education, UNESCO), and the public sector and employment (OECD, EU AI Act Annex III) (Section 7, Tables~\ref{tab:sectoral_guidance} and \ref{tab:framework_comparison}).

\item An explicit findings synthesis answering each review question, together
with a critical limitations analysis that identifies four persistent gaps and
five structural tensions, each converted into a requirement any accountable
deployment must satisfy (Section 8, Tables~\ref{tab:persistent_gaps} and \ref{tab:structural_tensions}).

\item A consolidation of the surveyed evidence into an integrated accountability architecture addressing the identified gaps, with three cross-cutting properties and cybersecurity aligned to the OWASP LLM Top 10 (2025), presented as a synthesis of the corpus rather than a validated artefact (Section 9).

\end{itemize}

\subsection{Paper Organisation}

The remainder of the paper is organised as follows. Section 2 provides background and Section 3 describes the PRISMA-inspired methodology. Section 4 reports the review process and the characteristics of the corpus. Sections 5 to 7 survey the conceptual foundations, the mechanisms, and the regulatory landscape in turn. Section 8 consolidates the findings against the four review questions and reports the persistent gaps and structural tensions. Section 9 presents LAAF as the synthesis of that evidence, Section 10 sets out the research agenda, Section 11 discusses limitations and threats to validity, and Section 12 concludes. Readers concerned with mechanisms may begin at Section 6, with compliance at Sections 7 and 9.4, and with the architecture at Section 9.

\section{Background}

\subsection{Evolution of Large Language Models}

\begin{figure}[!h]
    \centering
    \includegraphics[width=0.8\linewidth]{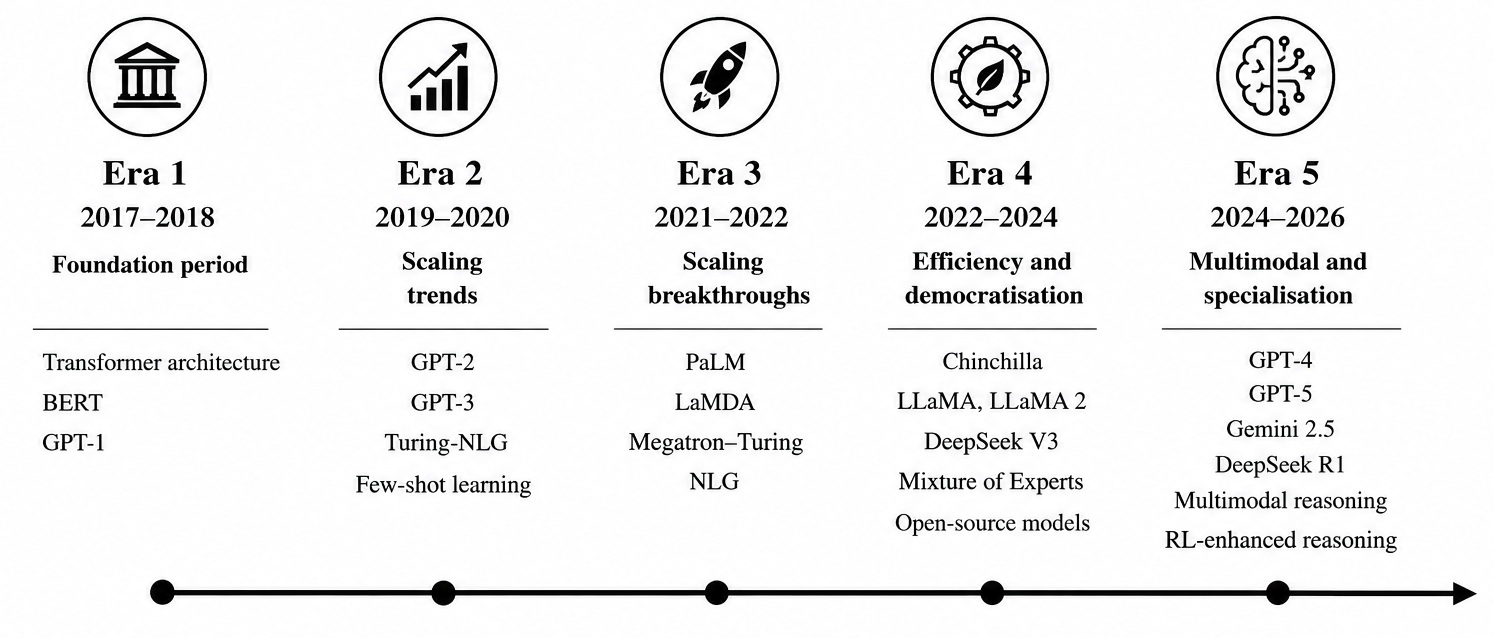}
    \caption{Evolution of Large Language Models, 2017--2026, in five eras: foundation (2017--2018), scaling (2019--2020), scaling breakthroughs (2021--2022), efficiency and democratisation (2022--2024), and multimodal and reasoning specialisation (2024--2026). The periodisation is the authors' construction, derived from the release history of the models named.}

    \label{fig:F1_LLM}
\end{figure}

LLMs are probabilistic models that factorise the likelihood of a token sequence into a product of conditional next-token probability \cite{huang2025survey, zhao2026survey}. Their distinguishing feature is scale: billions of parameters trained on heterogeneous corpora, giving rise to emergent abilities including reasoning, summarisation, translation, and conversation \cite{dominguez2024mapping, zhao2026survey}. Modern systems such as GPT, Claude, Gemini, and LLaMA are built on the transformer architecture \cite{vaswani2017attention},

combining self-supervised pre-training with supervised and reinforcement-learning-based fine-tuning \cite{zhao2026survey}. Figure \ref{fig:F1_LLM} presents a five-era timeline of this evolution.

The trajectory has moved from raw scale toward compute-optimal efficiency and reasoning \cite{zhao2026survey}. Two properties of this development matter for accountability. First, capability is general rather than task-bounded: outputs rest on statistical regularities rather than validated knowledge, producing answers that are plausible but incorrect and reproducing social and cultural biases, while the internal mechanisms that generate them remain opaque and raise privacy, intellectual-property, responsibility, and equity concerns \cite{das2025security, edenberg2023disambiguating}. Second, deployed behaviour is dynamic rather than fixed: providers update weights, safety layers, and system prompts after release, and deployers alter retrieval corpora and prompt logic, so the system a regulator inspects is not necessarily the system that runs the following week \cite{leon2026lifecycle,weidinger2023sociotechnical}. Accountability must therefore attach to a moving object, a point Section 2.5 develops and Section 7.4 revisits under predetermined change control.

\subsection{Taxonomy of LLM Harms}

\begin{figure}[!h]
    \centering
    \includegraphics[width=0.9\linewidth]{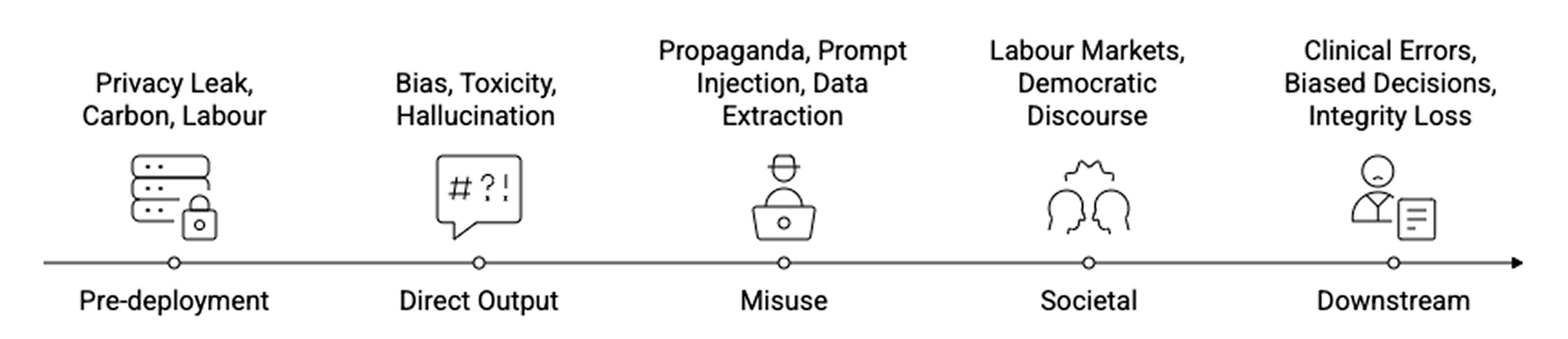}
    \caption{Taxonomy of LLM harms grouped by the lifecycle stage at which they originate. Arrows indicate that the phases are interdependent, an unaddressed harm at one stage resurfacing at the next; harms accumulate across the lifecycle, which is why accountability must be a lifecycle property. } 
    
    \label{fig:F2_LLM}
\end{figure}

The harms associated with LLMs extend across the entire lifecycle and arise from interactions among data practices, model architectures, deployment contexts, and socioeconomic systems \cite{dominguez2024mapping, weidinger2022taxonomy}. Existing taxonomies organise harms by type; because this survey is concerned with where accountability must attach, we group them instead by the lifecycle stage at which they originate, as shown in Figure~\ref{fig:F2_LLM}.

Pre-deployment harms originate in the construction of the model and comprise training-data privacy leakage and memorisation \cite{petrov2024dager}, environmental cost \cite{ren2024reconciling,liu2024green}, and exploitative annotation labour \cite{dominguez2024mapping}. Direct-output harms are properties of what the system generates: representational bias, toxicity, misleading content, and hallucination \cite{gallegos2024bias, huang2025survey}. Misuse and malicious-use harms arise when the system is turned to hostile purposes, including extremist propaganda and fraud \cite{hasanain2024large, hagendorff2024mapping}, prompt injection \cite{greshake2023not}, and training-data extraction \cite{petrov2024dager}. Societal and systemic harms operate at population scale, reshaping labour markets and democratic discourse \cite{dominguez2024mapping,hagendorff2024mapping}. Downstream application harms occur where LLM output displaces professional judgement in high-stakes contexts, producing clinical errors, biased decisions, and losses of institutional integrity \cite{dahl2024large, li2024chatgpt, omar2025sociodemographic}.

The five phases are interdependent rather than sequential: harms accumulate across the lifecycle rather than at a single point, the first argument for treating accountability as a lifecycle property.

\subsection{Hallucinations in LLMs}

A hallucination is an output that is fluent and well formed but not supported by the model's input, its retrieved context, or the world \cite{huang2025survey,wang2025survey}. The defining property is not confident delivery, since a model may hedge on a supported claim and assert an unsupported one, but that delivery carries no reliable signal of evidential support; the forms that mismatch takes are captured by the taxonomy below.

Coding the hallucination-focused studies yields five recurring types: factual contradiction (demonstrably false), contextual irrelevance (accurate but unresponsive), logical inconsistency (self-contradiction or constraint violation), temporal disorientation (events misplaced relative to the knowledge boundary), and ethical violation (breach of a stated norm). This classification is the authors' synthesis and is descriptive rather than quantitative: the studies use incompatible definitions, prompts, and domains, so no defensible pooled frequency can be assigned, itself an instance of the missing-metrics gap of Section 8.5.

These five types classify one direct-output harm and should not be confused with the five lifecycle phases of Section 2.2, in which hallucination is a single phase.

\begin{figure}[!h]
    \centering
    \includegraphics[width=0.9\linewidth]{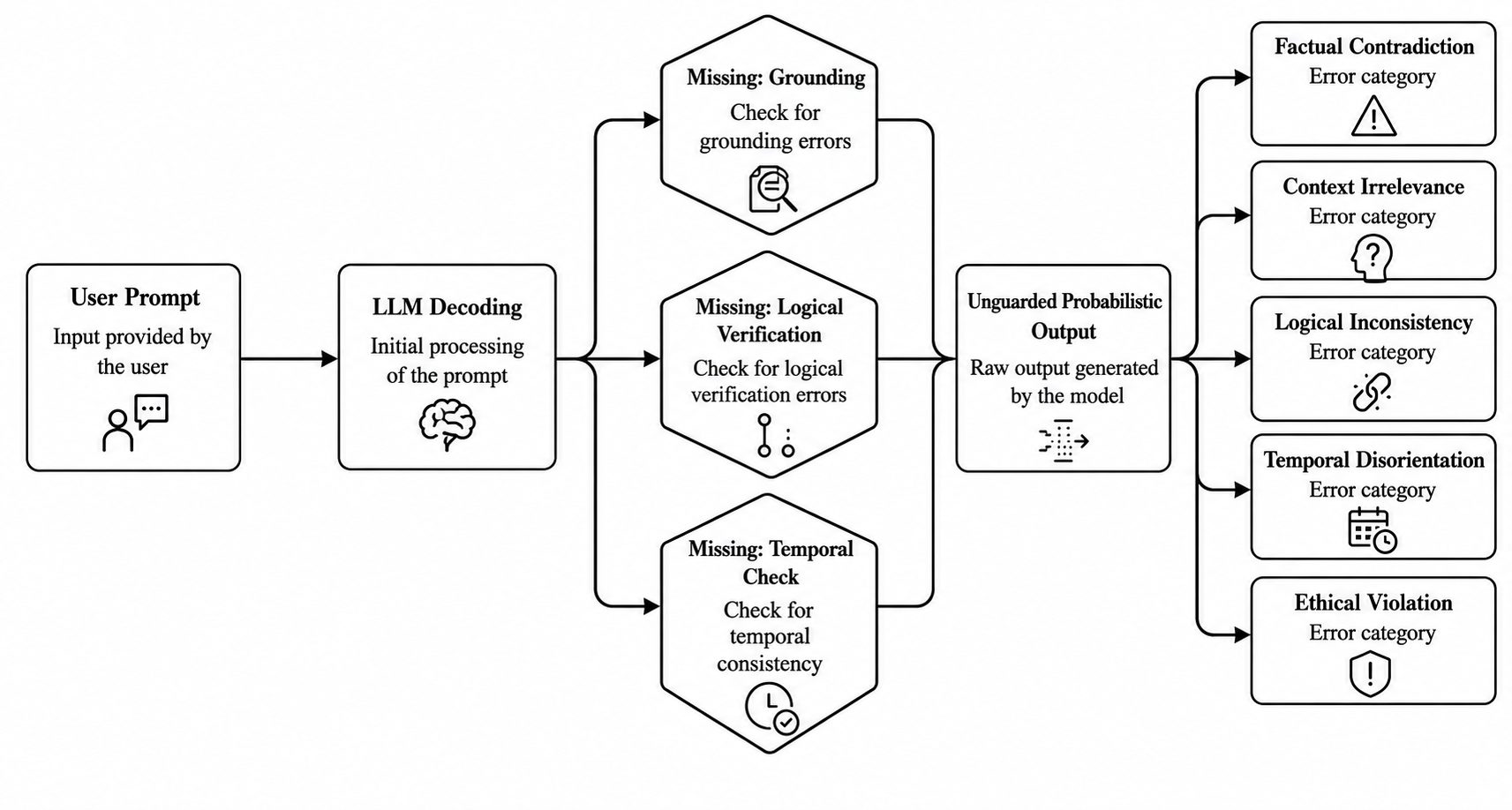}
    \caption{Emergence of hallucinations in LLM pipelines. Absent grounding, logical verification, and temporal consistency checking permit unguarded probabilistic decoding to produce the five recurring categories of hallucinated output.}
    
    \label{fig:F3_LLM}
\end{figure}

Figure \ref{fig:F3_LLM} shows how these outcomes arise in the generation pipeline, tracing the structural causes, namely absent grounding, absent logical verification, and absent temporal consistency checking, through unguarded probabilistic decoding to the five recurring categories of hallucinated output.

\subsection{Formalisms for Hallucination Detection}
\label{sec:formalisms} %label added by PD

Hallucination detection is formalised within a probabilistic sequence-generation
framework ~\cite{huang2025survey,wang2025survey}. The modelling assumption is that an autoregressive model
$f$ defines the conditional distribution of the next token,

\begin{equation}
P(x_t \mid x_{1:t-1}) 
\end{equation}

where $x_{1:t-1}$ is the prefix context~\cite{huang2025survey}. Equation (1) is not itself a
detection criterion; it is the generative process whose outputs the criteria in
Equations (2) to (4) examine.

Given a prompt $x_p$ and a response $x$, detection on a single response is
framed as binary classification,

\begin{equation}
H(x_p,x)\in\{0,1\}
\end{equation}

where $1$ indicates a hallucination ~\cite{manakul2023selfcheckgpt,huang2025survey}. Population-based methods
instead draw $k$ independent responses to the same prompt and analyse their
internal representations. Collecting the $d$-dimensional hidden states of the
$k$ sampled responses into a matrix

\begin{equation}
 Z \in \mathbb{R}^{k\times d},
\end{equation}

with $\bar{Z}$ the row-wise mean, the sample covariance is

\[
\Sigma = \frac{1}{k-1}(Z-\bar{Z})^{T}(Z-\bar{Z})
\]

A broadly dispersed spectrum indicates low agreement across the $k$ samples and is taken as sample-level evidence of hallucination; the same signal is obtainable without access to model internals, whether by entropy over clusters of equivalent meanings  ~\cite{farquhar2024detecting} or by pairwise consistency between sampled answers \cite{manakul2023selfcheckgpt}.

When external references
\[
R=\{r_1,\ldots,r_l\},
\]
are available, detection compares the response against them. Let

\[
   g(x,R)\in[0,1] 
\]

be an alignment score measuring how far the claims in $x$ are entailed by $R$ (via natural-language inference or retrieval similarity). A hallucination is flagged when

\begin{equation}
g(x,R)<\tau
\end{equation}

as in retrieval-augmented detection~\cite{mishra2024fine, niu2024ragtruth}. The threshold $\tau$ is not a universal constant. It is calibrated on a labelled validation set for the specific deployment, trading false alarms against missed hallucinations according to the relative cost of each in that setting. Choosing $\tau$ is therefore not only a technical decision but an accountability decision, and Section~9 assigns it to Layer~2 with an explicit record of the justification.

\begin{figure}[!h]
    \centering
    \includegraphics[width=1.0\linewidth]{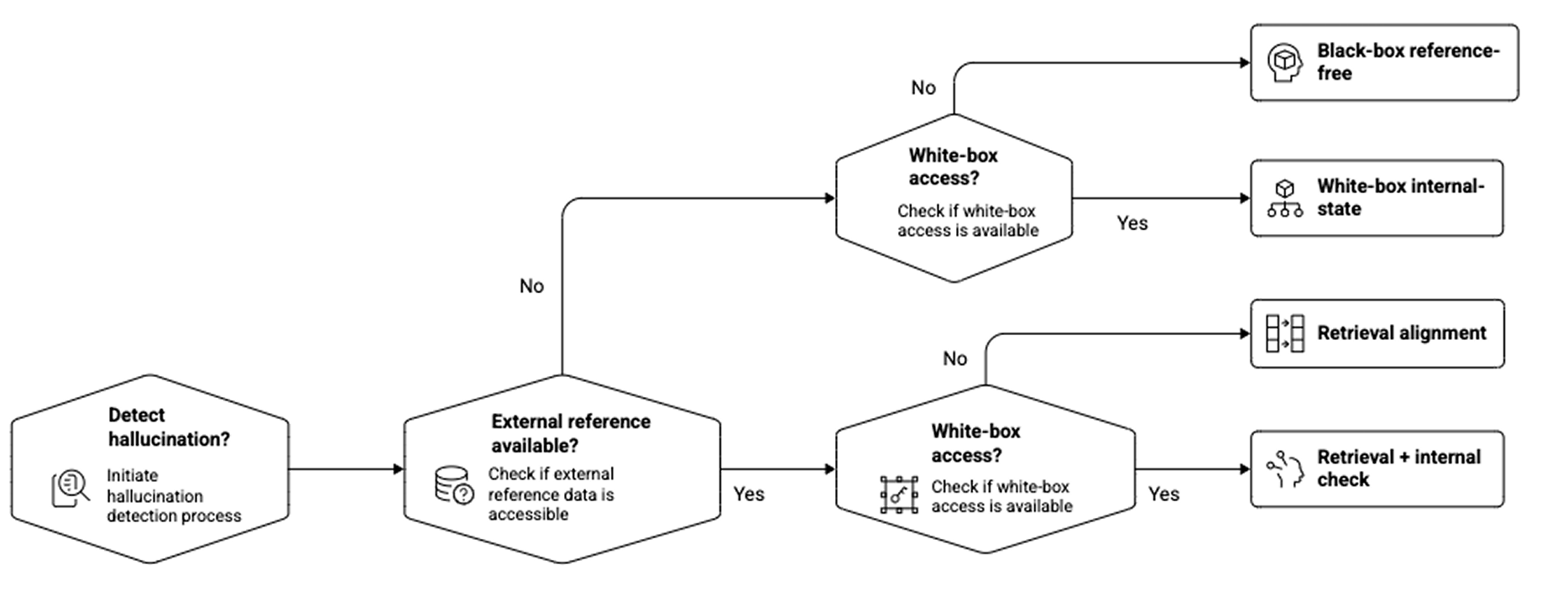}
    \caption{Hallucination-detection decision tree. Branching on two conditions, whether external reference material is available and whether the deployer has white-box access to internal model states, yields four method families: black-box reference-free, white-box internal-state, retrieval alignment, and retrieval combined with an internal check.}

    \label{fig:F4_LLM}
\end{figure}

Figure \ref{fig:F4_LLM} maps these to four procedures, branching on external-reference availability and model access (Equations (1) to (4)).

A worked case makes the two signals concrete. If ten sampled responses to a clinical query fall into meaning-clusters of 7, 2, and 1, the cluster entropy is about 0.80 nats (normalised dispersion $\approx 0.73$), which is sample-level evidence of hallucination without any reference corpus; if a guideline corpus is available and only three of four claims are entailed ($g(x,R)=0.75$) against a deployment threshold $\tau=0.85$, the output is flagged under Equation~(4) and routed to review. The dispersion signal measures internal uncertainty and is available even behind a closed API but cannot catch confident, consistent errors, whereas the alignment signal measures support by an authority but presupposes a corpus that encodes the standard.

These formalisms matter because answerability presupposes detectability: an organisation cannot be required to explain, correct, or redress an output it has no means of identifying as erroneous, and each detection regime corresponds to a different level of assurance a deployer can offer a forum. Hallucinations are consequently not occasional malfunctions but a systemic property of probabilistic generation \cite{huang2025survey,wang2025survey}, and any account of accountability must begin by acknowledging this.

\subsection{The Sociotechnical Nature of LLM Deployments}
A common misconception treats the model as the object of governance, but this does not survive contact with how LLMs are actually deployed \cite{nabben2024ai,weidinger2023sociotechnical}. An LLM sits at the centre of a layered arrangement of four actor groups: providers, who supply base weights, alignment, and release decisions; integrators, who build retrieval pipelines, prompt logic, tools, and interfaces on top of them; deployers, who set policy, allocate roles, and place the system in a workflow; and overseers, comprising internal reviewers and external regulators. Table~\ref{tab:actors_accountability} gives each group's contribution and the artefact through which it can be held answerable.

No single actor holds a complete view of the system's behaviour or risk, which is why scholarship increasingly treats LLM applications as constituted systems rather than artefacts, meaning that the object of governance is the assembled arrangement of models, data, interfaces, and institutional practice rather than the model in isolation \cite{nabben2024ai,weidinger2023sociotechnical}. Two consequences follow, and they are parallel in form. Accountability must be a lifecycle property, maintained through documentation, monitoring, and incident review across the whole period of deployment \cite{leon2026lifecycle}. Accountability must equally be a distributed property, allocated across named actors in a way that preserves clarity over who decided what \cite{novelli2024accountability,green2022flaws}.

This framing sets the review's inclusion criteria (Section 3.3) and dictates the layered classification device of Section 2.6.

\begin{table*}[!h]
\centering
\caption{Actors, contributions, and accountability surfaces in LLM deployments. Identifies each actor's contribution and the artefact through which they can be held answerable; the takeaway is that accountability is a relation across actors, not a property of the model. The regulator is listed as an external actor whose entry in the final column records the instruments through which it holds others answerable, rather than an artefact for which it is itself answerable. The final column names the LAAF layer (Section 9.2) at which each actor becomes answerable; the affected party and the regulator are forums to which the stack answers, not positions within it.}
\label{tab:actors_accountability}
\resizebox{\textwidth}{!}{%
\begin{tabular}{p{1.6cm} p{3.0cm} p{4.4cm} p{5.0cm} p{2.0cm}}
\hline
\textbf{Actor type} & \textbf{Actor} & \textbf{Contribution} & \textbf{Accountability surface} & \textbf{Answerable at}  \\
\hline

Internal &
Foundation-model provider &
Training data, base weights, alignment, release &
Model card, evaluation reports, systemic-risk disclosures & L1 \\

Internal &
Fine-tuning party &
Domain adaptation, safety patches &
Fine-tuning records, drift documentation & L1 \\

Internal &
Application developer &
Prompt design, retrieval, tools, guardrails &
Application logs, design rationale, retrieval provenance & L2\\

Internal &
Deploying organisation &
Policy, roles, review obligations, override authority &
Governance policy, role allocation, audit trails & L2-L4\\

Internal &
Domain professional &
Prompt phrasing, interpretation, professional judgement &
Review records, sign-off, duty of care & L3\\

External &
End user or affected party &
Reliance on output, downstream action &
Terms of use, reported harms, complaint and contestation records & Forum, not a layer \\

External &
Regulator &
Risk classification, conformity assessment, monitoring &
Audit, sanctions, mandated reporting and remediation & External forum \\

\hline
\end{tabular}
}
\end{table*}

\subsection{A Layered Classification Device for the Survey}

Table~\ref{tab:actors_accountability} assigns each actor group an accountability surface, that
is, the artefact through which it can be held answerable. Those surfaces fall into four groups ordered by distance from the model. Closest to the artefact are the surfaces that establish what the model is and where it came from, which only the provider or fine-tuning party can produce. Next are the surfaces generated when the model is put to use in an application, which belong to whoever built the retrieval pipeline, prompt logic, and guardrails. Next are
the records of human review, which belong to the professional in the path of the output. Furthest from the model are the institutional surfaces, namely policy, audit, incident response, and redress, which belong to the deploying organisation and to the forums that hold it to account.

We adopt these four loci as the classification device through which the corpus is read: L1 provenance, L2 application logic, L3 human oversight, and L4 governance and redress. Three properties cut across them and serve as the axes for comparing studies within each layer: traceability, whether a mechanism leaves a record from which a decision can be reconstructed; role clarity, whether it resolves to a named answerable actor; and continuous monitoring, whether it operates beyond the point of deployment. The device is a taxonomy for organising a heterogeneous literature, not a design we propose. Its warrant is that it partitions the accountability relation by the question a forum actually asks, namely who can be required to explain this and on the basis of which artefact, which no organisation by technique can do. Section 5 surveys mechanisms layer by layer on this basis, and Section 9 consolidates the device into an integrated architecture.

\section{Methodology}

\subsection{Review Protocol and Research Questions}

The methodology follows a structured systematic-review protocol designed for transparency, reproducibility, and balanced coverage, drawing on PRISMA reporting guidance \cite{sohrabi2021prisma,page2021prisma}, and established evidence-based practice for systematic reviews in computer science \cite{dyba2005evidence,kitchenham2009systematic}.

The protocol, comprising the review questions, search strategy, inclusion and exclusion criteria, extraction template, and synthesis approach, was specified before any search was executed \cite{kitchenham2009systematic}. 

The protocol was not registered with PROSPERO or OSF, and we record this as a limitation in Section 11.

The protocol governs all four review questions. RQ1 to RQ3 direct attention respectively to conceptual treatment, mechanisms, and regulatory alignment, and are answered from the extracted evidence in Sections 5 to 7. RQ4 asks what patterns and gaps persist across that evidence; it is answered in Section 8 by reading the three bodies of evidence together, and Section 9 consolidates those findings into an integrated account.

\subsection{Search Strategy and Databases}

To ensure broad, high-quality coverage, the search was restricted to peer-reviewed sources indexed in five databases: the ACM Digital Library, IEEE Xplore, Elsevier ScienceDirect, SpringerLink, and Wiley Online Library, the last included for specialist journals at the intersection of AI and risk analysis \cite{arvanitou2021software,dyba2005evidence}. Preprint servers such as arXiv and the ACL Anthology were not screened systematically alongside the five databases. Preprints are instead admitted as a separate documented stratum under the stricter rule stated in Section 3.3, so that peer review remains a uniform quality floor for the 122 while coverage of the technical literature is not lost. Where a preprint has since appeared in a peer-reviewed venue, the published version was screened and counted in the main corpus. The treatment of the two strata is recorded as a limitation in Section 11.

Queries were executed against title, abstract, and author keywords. The search string combined three Boolean blocks, joined by AND, with terms within each block joined by OR:

("large language model" OR "LLM" OR "foundation model" OR "generative AI" OR "transformer-based model") AND ("accountability" OR "answerability" OR "oversight" OR "audit" OR "traceability" OR "governance" OR "risk management") AND ("healthcare" OR "finance" OR "law" OR "legal" OR "education" OR "public sector" OR "software engineering" OR "cybersecurity" OR "decision support")

All queries were executed on 31 March 2026, which is the search cutoff; the window therefore runs from 1 January 2022 to 31 March 2026. Searching proceeded in three rounds: the database keyword search above (4,512 records; the round the PRISMA flow reports); citation chasing, surfacing 38 candidates of which 9 met the criteria and are counted within the 122; and a purposive search for the 12 regulatory documents, analysed in Section 7 but not coded or counted among the 122.

\subsection{Inclusion and Exclusion Criteria}

A record was included if it was peer-reviewed, indexed in the selected databases, published between January 2022 and March 2026, addressed at least one dimension related to accountability, and written in English \cite{kitchenham2009systematic}. Authoritative regulatory and standards documents were added as primary sources rather than through database screening \cite{ai2023artificial,ai2024artificial,iso2023iso}. A record was excluded if it addressed only narrow technical performance, concerned pre-LLM systems, was a non-peer-reviewed source (e.g. blog or vendor whitepaper), or duplicated material already included.
Preprints were admitted as a separate documented stratum under a stricter rule: a preprint was included only where it introduces a formalism, benchmark, or result on which included peer-reviewed studies depend. Preprints are reported separately from the 122 and are excluded from the quality-score distribution of Section 4.3, since the reproducibility criterion presumes peer review.

\subsection{The PRISMA Flow}

The screening process followed three sequential stages: identification, screening, and eligibility illustrated in Figure 5 and reconciled with the per-database counts of Table \ref{tab:screening_results} in Section 4.

\begin{figure}
    \centering
    \includegraphics[width=0.7\linewidth]{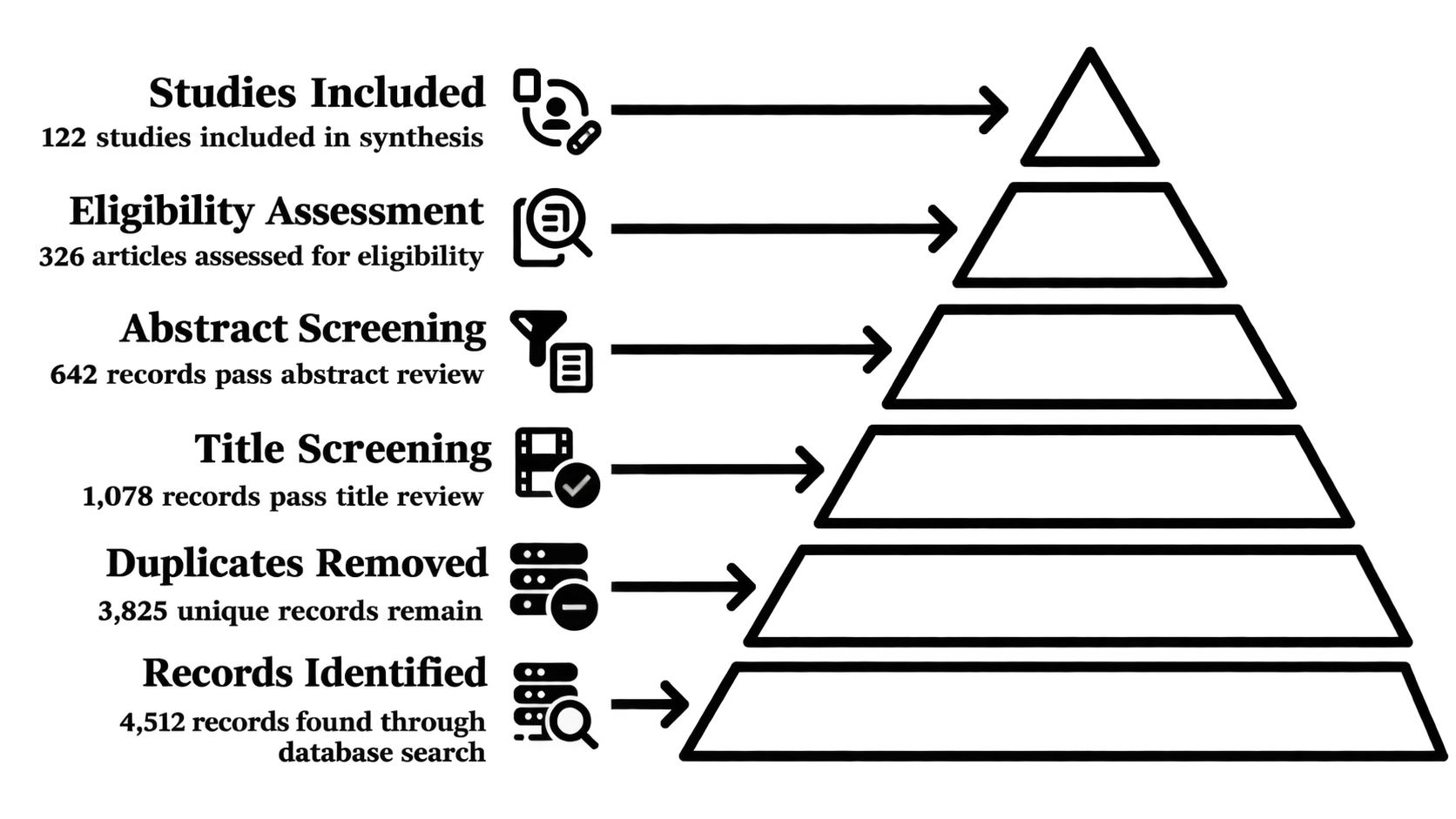}
    \caption{PRISMA 2020 flow of the review process, reporting exclusions at each stage, and reconciling with the per-database counts of Table \ref{tab:screening_results}. 
    % Twelve regulatory and standards documents were added purposively and are not counted among the 122.
    }

    \label{fig:F5_LLM}
\end{figure}

Figure \ref{fig:F5_LLM} reports the cascade: 4,512 identified, 3,825 after duplicate removal, then 1,078, 642, 326, and finally 122 included after successive title, abstract, introduction-and-conclusion, and full-text screens, with 12 regulatory documents added as primary sources.

\subsection{Data Extraction and Coding}
\label{sec:extraction} %label added by PD

A template was designed to collect bibliographic, contextual and content metadata related to the review questions \cite{xiao2019guidance}. It documented the Definition, Actors, Forums, and neighbouring concepts for RQ1; the Mechanism family, specific mechanisms, Maturity, Domain, and Limitations for RQ2; and the gaps and tensions that feed the framework for RQ3 and RQ4.

\subsection{Synthesis and Quality Assessment}
\label{sec:quality} %label added by PD

Synthesis combined narrative synthesis with thematic analysis following Braun and Clarke's six-phase procedure, which is appropriate for heterogeneous evidence not amenable to quantitative meta-analysis~\cite{braun2006using}. Themes were derived deductively against the four review questions at the first pass, then refined inductively where recurrent content did not fit an existing theme, with the resulting theme set applied to the full corpus on a second pass.

Each record was scored on three equally weighted criteria adapted from software-engineering review practice~\cite{xiao2019guidance,kitchenham2009systematic}: empirical rigour (3--1), reproducibility (3--1), and conceptual contribution (3--1); equal weighting avoids demoting the governance literature RQ1 and RQ3 depend on. Totals of 7--9 are high, 4--6 medium, and 3 conceptual-only, the last denoting unvalidated rather than low-value work.

Quality scores were used to weight the synthesis narratively, so that claims resting only on conceptual sources are marked as such in Sections 5 and 8; they were not used as an inclusion filter, since excluding conceptual work would remove much of the accountability literature the review exists to survey. Each record was scored against the prespecified rubric in two independent passes separated in time, and the two passes were reconciled by re-reading the source text before the scores were fixed; residual scoring subjectivity is recorded in Section 11 as a limitation. The score distribution is reported in Section 4.3. Threats were mitigated by combining five databases, applying prespecified criteria, acknowledging English-only inclusion as a limitation, and separating conceptual from mechanism extraction.

\subsection{Scoring Rubric for the Comparative Tables}

Table~\ref{tab:related_work} and Table~\ref{tab:laaf_comparison} are scored against an observable rubric. Y or $\checkmark$ denotes that the work treats the criterion as a structured component, evidenced by a dedicated section, table, or framework element addressing it; P that the criterion is discussed in prose without a dedicated structural element; N or $\times$ that it is not addressed. Scoring was performed on the full text of each work, and a criterion appearing only in a future-work section was scored N. Both tables are single-rater scored against the same rubric.

The two tables serve different purposes: Table~\ref{tab:related_work} compares this survey against related surveys, that is, what the literature covers, while Table~\ref{tab:laaf_comparison} compares the architecture of Section 9 against governance frameworks, that is, what frameworks prescribe. Several entries appear in both, scored on different criteria.

\begin{table*}[!h]
\centering
\caption{Comparative analysis of related surveys and contributions. Each prior work is scored on five comparison criteria: Concept, Mechanisms, Regulatory mapping, Sector cases, Cybersecurity, plus its differentiation; the dense pattern demonstrates the integration gap. (Y = covered, P = partial, N = absent), Y denotes that the work treats the criterion as a structured component with a dedicated section, table, or framework element; P that it is discussed without being developed; N that it is absent.}

\label{tab:related_work}
\resizebox{\textwidth}{!}{%
\begin{tabular}{p{0.8cm} p{2.3cm} c c c c c p{5.5cm}}
\hline
\textbf{Ref.} & \textbf{Focus} & \textbf{Concept} & \textbf{Mech.} & \textbf{Reg.} & \textbf{Sector} & \textbf{Cyber} & \textbf{Differentiation from this survey} \\
\hline

\cite{huang2025survey} & Hallucination taxonomy & P & Y & N & N & N & We integrate hallucination into an accountability framework \\

\cite{novelli2024accountability} & Accountability concept & Y & N & N & N & N & We add mechanism synthesis + LAAF construction \\

\cite{zeiser2024owning} & Attributability gap & Y & N & N & N & N & We extend to LLMs with regulatory mapping \\

\cite{mokander2024auditing} & LLM auditing (3-layer) & P & Y & P & N & N & LAAF covers full lifecycle incl. cybersecurity \\

\cite{hacker2023regulating} & EU AI Act analysis & P & N & Y & P & P & We bridge legal and technical perspectives \\

\cite{schuett2025three} & Risk governance for AI & P & P & P & N & N & We update to LLM context with GenAI Profile \\

\cite{gallegos2024bias} & LLM bias \& fairness & P & Y & N & P & N & We treat bias within broader accountability \\

\cite{bommasani2023foundation} & Transparency Index & P & P & N & N & N & We position transparency within accountability layers \\

\cite{da2024survey} & LLM cybersecurity survey

& P & P & P & N & Y & We construct an integrative framework 
 
\\

\cite{edenberg2023disambiguating} & Algorithmic-bias disambiguation 

& P & N & Y & P & N & We integrate regulation with technical mechanisms \\

\cite{naser2025auditing} & Shared training-data auditing 
& P & Y & P & N & N & We add regulatory + sector + cybersecurity integration \\

\cite{das2025security}  & LLM security \& privacy    & P & Y & N & N & Y & We situate security within the accountability relation \\

\cite{wang2025survey} & Factuality survey           & P & Y & N & N & N &  We extend from output properties to answerability \\

\cite{ferdaus2026towards}  & Trustworthy \& robust LLMs & Y & P & P & N & P &  We map principles onto binding obligations \\

\textbf{Ours} & \textbf{LAAF framework} & \textbf{Y} & \textbf{Y} & \textbf{Y} & \textbf{Y} & \textbf{Y} & To our knowledge, the only work integrating all five criteria; scoring is author-assigned and reported with its rubric in Section 3.7.

\\

\hline
\end{tabular}
}
\end{table*}

\section{Review Process Reporting and Dataset Characteristics}
\label{sec:reviewprocess}

This section reports the review process quantitatively before the synthesis begins, so that the evidence base is visible in advance of the claims drawn from it; the search, screening, and adjudication procedures are those of Sections 3.2 and 3.6. Records were exported from each database, deduplicated in Zotero, and screened in a shared spreadsheet. The figures here are descriptive; Sections 5 to 8 synthesise the corpus they characterise.

\subsection{Review Process Outcome}

Table \ref{tab:screening_results} reports per-database screening at each stage. The final 122 studies represent 2.7\% of the 4,512 records identified. 

Records appearing in multiple databases were retained once, deduplicated by Digital Object Identifier; where a record carried no DOI, deduplication was by exact title and first-author match. The nine studies surfaced by citation chasing are attributed in Table \ref{tab:screening_results} to the database indexing them, so the per-database totals sum to 122; they are flagged in Table \ref{tab:screening_results}.

The 12 regulatory documents came from the purposive round-three search (Section 3.2); they were not coded or counted among the 122, and are analysed in Section 7 and used as the mapping targets in Section 9.4. 

The cascade adds one stage beyond standard PRISMA: after abstract screening, records were assessed on introduction and conclusion, because governance abstracts often signal relevance without indicating substantive treatment, and full-text assessment of 642 records was not proportionate; the stage is reported distinctly in Table \ref{tab:screening_results}.

\subsection{Characteristics of Included Studies}

The 122 studies show methodological diversity: conceptual or survey 39 (32\%), empirical case studies 28 (23\%), technical mechanism 33 (27\%), and regulatory or governance 22 (18\%). Temporal distribution favours recent work: 2022--2023, 31 (25\%); 2024, 30 (25\%); 2025, 45 (37\%); and January--March 2026, 16 (13\%), reflecting intensified attention as the EU AI Act approached enforcement.

\subsection{Quality Assessment Outcome}

Applying the framework in Section~3.6, 28 studies (23\%) scored high, 71 (58\%) scored medium, and 23 (19\%) scored conceptual-only. The predominance of medium-quality, largely simulated and conceptual work rather than real world deployment evidence is itself a key finding, and it motivates the empirical-evaluation gap discussed in Section 8.5.

\subsection{Benchmark Datasets and Evaluation Resources}
\label{sec:benchmarks}

Empirical evaluation of accountability mechanisms requires benchmarks. Table~\ref{tab:benchmark_datasets} summarises eight public resources covering hallucination, factuality, bias and fairness, and security

\cite{manakul2023selfcheckgpt,niu2024ragtruth}. These resources measure output level properties; none measures the oversight substantiveness, governance closure, or enforcement responsiveness identified in Section 5.3. The absence of benchmarks for these accountability dimensions rather than for output quality is the measurement gap examined in Section 10 \cite{yehudai2025survey,raji2020closing}.

\begin{table*}[!h]
\centering
\caption{Benchmark datasets and resources for accountability evaluation. Catalogues the principal public resources and their accountability use; the takeaway is that current benchmarks measure outputs, not accountability mechanisms.}
\label{tab:benchmark_datasets}
\resizebox{\textwidth}{!}{%
\begin{tabular}{p{3.0cm} p{2.5cm} p{3.8cm} p{4.0cm}}
\hline
\textbf{Dataset / resource} & \textbf{Domain} & \textbf{Scale} & \textbf{Application} \\
\hline

HaluEval \cite{li2023halueval} &
Hallucination detection &
35,000 annotated samples &
Evaluating detection methods \\

TruthfulQA \cite{lin2022truthfulqa} &
Factuality &
817 questions, 38 categories &
Truthful generation \\

FactScore \cite{min2023factscore} &
Long-form factuality &
Atomic-fact annotations &
Fine-grained factuality \\

RAGTruth \cite{niu2024ragtruth} &
RAG hallucination &
18,000 word-level annotations &
Retrieval-augmented evaluation \\

BOLD \cite{dhamala2021bold}&
Fairness &
23,679 prompts &
Open-ended fairness \\

HolisticBias \cite{smith2022m}&
Bias coverage &
600 descriptors, 13 axes &
Bias measurement / audit \\

OWASP LLM Top 10 (2025) \cite{shahin2026benchmarking} &

Security &
Ten categories, LLM01–LLM10 &
Red-teaming, security evaluation \\

BBQ (Bias Benchmark for QA) \cite{parrish2022bbq} &

Social bias in QA &
58,492 question templates &
Bias auditing across nine social dimensions \\

\hline
\end{tabular}
}
\end{table*}

\begin{table*}[!h]
\centering
\caption{Selection and screening results from digital libraries. The per-database PRISMA cascade; the takeaway is the progressive elimination from 4,512 to 122, reconciling exactly with Figure~5.}
\label{tab:screening_results}
\resizebox{\textwidth}{!}{%
\begin{tabular}{p{4.0cm} c c c c c c}
\hline
\textbf{Stage} & \textbf{ACM} & \textbf{IEEE} & \textbf{ScienceDirect} & \textbf{Springer} & \textbf{Wiley} & \textbf{Total} \\
\hline

Identified (keyword) & 612 & 1,847 & 1,108 & 832 & 113 & 4,512 \\

After duplicate removal & 553 & 1,648 & 989 & 565 & 70 & 3,825 \\

After title screening & 187 & 432 & 261 & 174 & 24 & 1,078 \\

After abstract screening & 92 & 248 & 178 & 109 & 15 & 642 \\

After intro/conclusion review & 45 & 121 & 89 & 64 & 7 & 326 \\

Final inclusion  & 24 & 41 & 32 & 22 & 3 & 122 \\

\hline
\end{tabular}
}
\end{table*}

\section{Conceptual Foundations of Accountability in the Literature}
\label{sec:rq1}
\subsection{Defining Accountability}

Accountability is among the most frequently invoked and most imprecisely used ideas in AI governance \cite{novelli2024accountability, wieringa2020account}. It is not a synonym for transparency, explainability, or responsibility. It is tied to answerability: identifying the relevant actors, requiring them to explain and justify their conduct against applicable standards, and subjecting them to oversight or consequences when failures occur \cite{green2022flaws}. This framing derives from Bovens's account of public accountability as a relation between an actor and a forum, in which the actor is obliged to explain and justify conduct, the forum may pose questions and pass judgement, and consequences may follow \cite{bovens2007analysing}. It is sharper than the alternatives because it specifies not only that some actor must be accountable, but to whom, on what grounds, against what standards, and with what consequences \cite{novelli2024accountability}.

The definition below consolidates that convergence for the sociotechnical setting of Section 2.5.

Definition (accountability in LLM applications). The structured capacity of a sociotechnical system to identify responsible actors across the lifecycle, require them to justify design and deployment choices, preserve traceability over how outputs are produced and used, and enable oversight, correction, and redress when harms occur.

This definition keeps accountability broader than model interpretability alone \cite{doshi2017towards, hassija2024interpreting}. Figure 6 visualises the structure. Answerability names the whole relation; the five dimensions of Section 5.3 are its components, the first of which is disclosure and justification rather than answerability itself.

\begin{figure}[!h]
    \centering
    \includegraphics[width=0.8\linewidth]{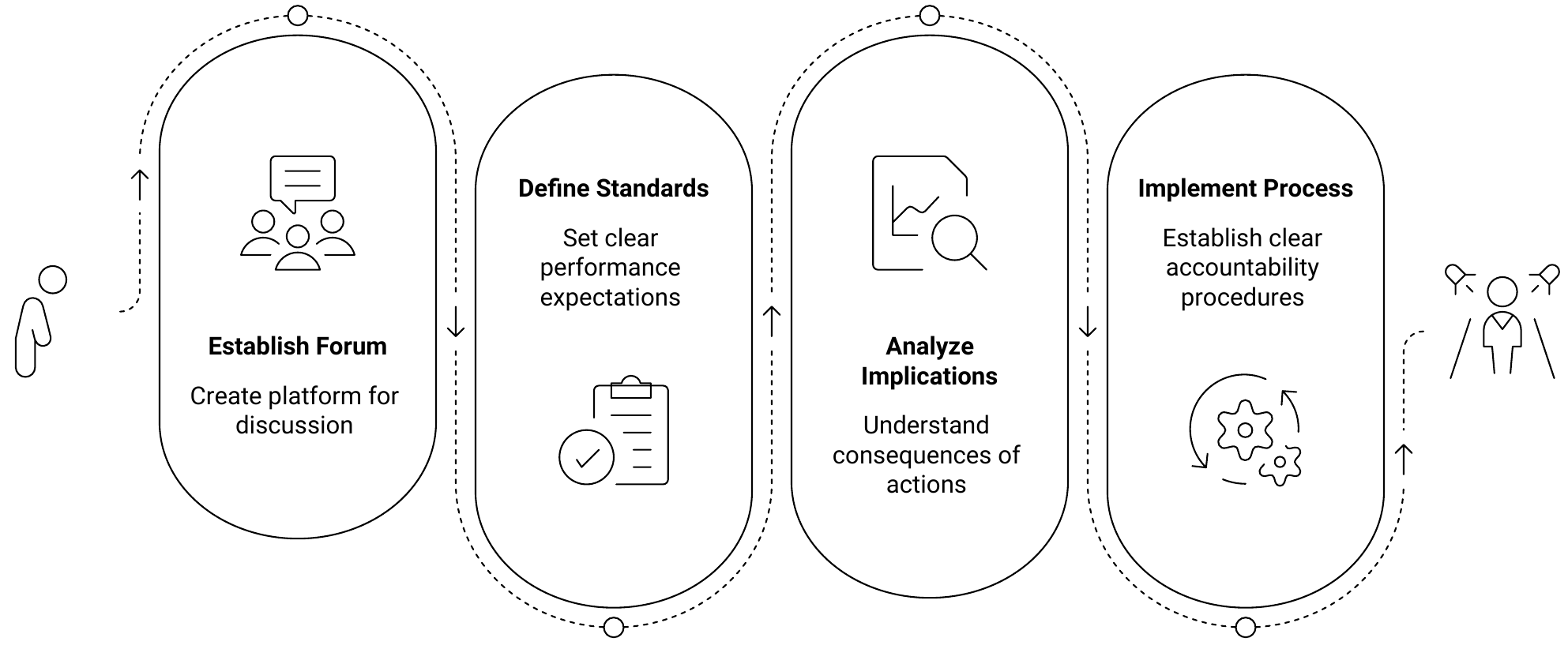}
    \caption{Conceptual structure of accountability in LLM applications. An actor is answerable to a forum, is judged against standards, explains conduct through a defined process, and faces implications; the relation holds only when all five elements are present.}
    \label{fig:F6_LLM}
\end{figure}

Figure \ref{fig:F6_LLM} presents the answerability relation in terms of five interdependent elements: the actor whose conduct is in question, the forum to which the actor is answerable, the standards against which the actor's conduct will be judged, the process in which the actor will be called for explanation, and the implications that will follow.

\subsection{Why LLM Applications Complicate Accountability}

Three interrelated aspects complicate accountability. First, the attributability gap: faced with confident output, a human may defer to the model rather than substantively own the decision, weakening the link to an answerable actor \cite{zeiser2024owning,vasconcelos2023explanations}.

Second, multi-layered dependencies: developers depend on closed foundation models, third-party interfaces, retrieval pipelines, and moderation layers, so no single point holds responsibility

\cite{nabben2024ai,jiao2025generative}. Third, dual-phase generative accountability: the interaction occurs first between the affected user and the generative system, with the human actor receding into a second phase, so the system appears responsive while the locus of human responsibility is obscured \cite{elliott2025evolving,zeiser2024owning}.

\subsection{Five Dimensions of Accountability}
\label{sec:dimensions} %label added by PD

The literature indicates accountability is not monolithic but comprises multiple aspects \cite{wieringa2020account}.

Five analytically distinct dimensions emerge. Disclosure and justification consists of questions regarding decisions in model selection, data, prompts, retrieval, evaluation and the known limitations of models \cite{green2022flaws}. Oversight asks whether human actors meaningfully review, challenge, override, and escalate \cite{sterz2024quest,asadollahi2026governing}. Organisational and governance asks whether roles and escalation pathways are embedded institutionally \cite{schuett2025three,van2026agentic}. Moral and professional asks whether responsibility remains attached to human actors rather than displaced onto the system \cite{constantinescu2021understanding}.

Enforcement and redress ask whether failures are investigated, corrected, sanctioned, and redressed \cite{jongepier2022explanation,raji2020closing}. Without the last dimension, accountability risks becoming purely rhetorical \cite{novelli2024accountability}. Coding the corpus for which dimensions each study substantively treats shows an uneven distribution: oversight and disclosure dominate, while enforcement and redress is treated least often, the conceptual counterpart of the missing-metrics gap of Section 8.5. Table~\ref{tab:five_dimensions} summarises the dimensions and failure modes; they reinforce one another, so a system can satisfy one or two while remaining unaccountable overall.

\begin{table*}[!h]
\centering
\caption{Five dimensions of accountability and failure modes. States the central question each dimension answers and what breaks when it is missing; the takeaway is that accountability emerges only from the interaction of all five.}
\label{tab:five_dimensions}
\begin{tabular}{p{3.2cm} p{5.5cm} p{5.0cm}}
\hline
\textbf{Dimension} & \textbf{Central question} & \textbf{Failure mode when missing} \\
\hline

Disclosure and justification &
Can actors explain and justify decisions? &
Disclosure without consequence \\

Oversight &
Can humans meaningfully review and override? &
Ceremonial sign-off, automation bias \\

Organisational / governance &
Are roles and pathways embedded institutionally? &
Aspiration without operationalisation \\

Moral/professional &
Are human duties of care preserved? &
Responsibility displaced onto the model \\

Enforcement/redress &
Are failures investigated and sanctioned? &
Rhetorical accountability without consequence \\

\hline
\end{tabular}
\end{table*}

\subsection{Implications for the Running Examples}

Three deployments recur as the running examples: a clinician reading an LLM-generated clinical summary, a credit officer reviewing a credit memo, and a public servant drafting a policy note. Each was sketched in Section 1.1.

The definition has parallel implications in all three: a named professional retains the duty of care or responsibility for the decision, the output is traceable to its sources or input factors, and the affected party retains a route to contest the result: the patient through clinical governance \cite{asadollahi2026governing}, the customer under CFPB Circular 2023-03 \cite{bureau2022consumer}, and the citizen through administrative contestation \cite{jongepier2022explanation}.

The five dimensions organise what follows: answerability and traceability map to Sections 6.1 and 6.2, oversight to Section 6.3, and the organisational, moral, and enforcement dimensions to Section 6.4.

\section{Accountability Mechanisms: A Systematic Synthesis}
\label{sec:rq2}

Mechanisms are surveyed by the layer at which they attach, following the classification device of Section 2.6, and cross-tabulated against four intervention families: technical controls, human oversight, organisational governance, and documentation and traceability. The two axes are not redundant: documentation mechanisms appear at every layer, but a model card is answerable by the provider whereas an incident record is answerable by the deploying organisation, and reading them as one family obscures that the two fail for different reasons and are fixed by different actors.

\begin{figure}[!h]
    \centering
    \includegraphics[width=0.8\linewidth]{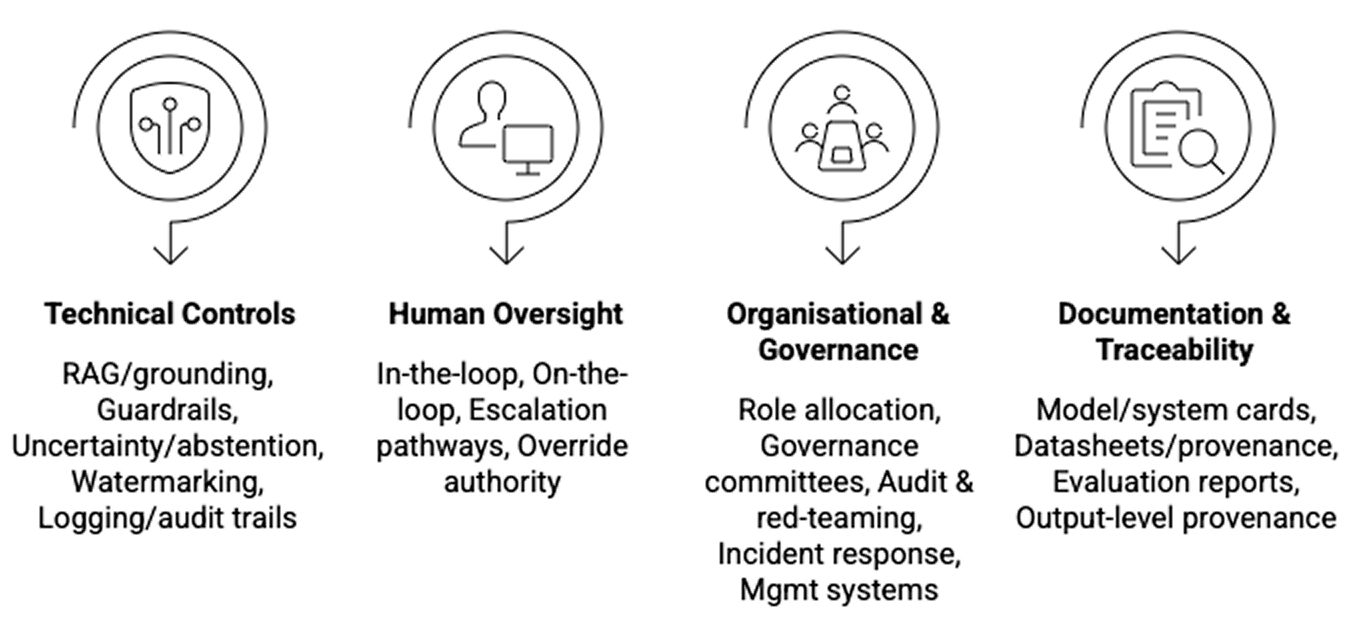}
    \caption{Taxonomy of accountability mechanisms for LLM applications: four families, each subdividing into specific mechanisms; no family suffices alone. }
    \label{fig:F7_LLM}
\end{figure}

Figure \ref{fig:F7_LLM} shows the four intervention families and the mechanisms within each; the layer at which each attaches is given in Table~\ref{tab:mechanism_families}. The point of both is that no single family or layer
suffices alone.

\subsection{Layer 1: Provenance Mechanisms}
\label{sec:layer1}
Four mechanisms dominate: model and system cards \cite{mitchell2019model}, datasheets and data provenance \cite{gebru2021datasheets}, evaluation and red-team reports \cite{ganguli2022red}, and watermarking, which embeds detectable signals in generated text for post-hoc attribution \cite{kirchenbauer2023watermark, zhao2023provable}.

Maturity is split: cards and datasheets are high; watermarking is emerging,
with no standardised scheme in production use. The principal limitation at this layer is that documentation records what was intended rather than what occurred, so a model card describes a release while the deployed system may have diverged from it through fine-tuning, retrieval changes, or prompt updates.

\subsection{Layer 2: Application-Logic Mechanisms}
\label{sec:layer2} %label added by PD
Grounding and retrieval is the most widely adopted control, RAG constraining generation to retrieved documents with significant hallucination reductions \cite{gao2023retrieval,izacard2023atlas}. Guardrails filter inputs and check outputs \cite{inan2024llama, rebedea2023nemo, jalan2026survey};

uncertainty quantification and abstention surface low-confidence outputs \cite{kuhn2023semantic, lin2022teaching}; logging and audit trails provide the retrospective evidence base mandated by the EU AI Act and ISO/IEC 42001 \cite{iso2023iso,lognoul2025regulation}. Output-level provenance records model version, prompt, retrieval set,
configuration, and uncertainty per output, which regulators increasingly expect
of high-risk systems \cite{wang2026agent, sachan2024blockchain}.

Maturity is uneven: RAG and logging are high, guardrails and uncertainty
quantification medium, and output-level provenance emerging, with no standardised schema (Table~\ref{tab:mechanism_families}). The principal limitation is that every mechanism operates on the surface of the output rather than the reasoning behind it: a policy-compliant hallucination passes a guardrail, a confident error passes an uncertainty threshold, and a log records production without establishing correctness.

\subsection{Layer 3: Oversight Mechanisms}
\label{sec:layer3} %label added by PD
The central distinction is between ceremonial review, formal sign-off and substantive review, which involves active interrogation of evidence, uncertainty, and reasoning \cite{sterz2024quest,asadollahi2026governing}. Four forms recur: human-in-the-loop review places a reviewer in the path of each output \cite{sharma2026human}; human-on-the-loop monitoring places a supervisor over aggregate behaviour \cite{singh2026architecting}; escalation pathways specify triggers, named parties, and response times \cite{polemi2023multilayer}; and override authority is the formal right to refuse, modify, or replace an output \cite{vasconcelos2023explanations}. The literature warns repeatedly that in-the-loop review degenerates into ceremonial sign-off when reviewers are overloaded or output is sufficiently fluent, so substantive review requires evidence, uncertainty information, retrieval citations, and clear criteria \cite{asadollahi2026governing,bansal2021does}.

Maturity is medium: widely specified and partially tooled, but independently evaluated mostly under simulated conditions. Effectiveness at this layer depends on what it cannot guarantee: reviewer capacity, expertise, and institutional willingness to bear the cost of an override.

\subsection{Layer 4: Governance and Redress Mechanisms}
\label{sec:layer4} %label added by PD

Five mechanisms dominate:
role allocation \cite{van2026agentic}, governance committees \cite{hadley2025investigating}, internal and red teaming \cite{ganguli2022red}, incident response and post-deployment monitoring \cite{yehudai2025survey}, and formal AI management systems \cite{mateo2026reference}. The publication of ISO/IEC 42001 in 2023 created the first international standard for an AI management system, signalling organisational commitment to embed accountability into operations \cite{iso2023iso}.

Maturity is medium: established in adjacent domains and codified in a certifiable standard, but with scarce AI-specific evaluation. The principal limitation at this layer is that organisational structures can be constituted without being consequential, so a committee may exist and convene without any recorded instance of it altering a deployment decision.

\begin{table*}[!h]
\centering
\caption{Accountability mechanism families (consolidated). Each family's primary contribution, maturity, and regulatory anchor; the takeaway is complementarity, each family serves different accountability dimensions, so LAAF integrates rather than privileges any one. Maturity is assigned from the three signal counts: high where all three are well represented, medium where two are, and emerging otherwise.}
\label{tab:mechanism_families}
\resizebox{\textwidth}{!}{%
\begin{tabular}{p{2.2cm} p{3.8cm} p{3.2cm} p{3.2cm} p{2.8cm}}
\hline
\textbf{Family} & \textbf{Specific mechanisms} & \textbf{Primary contribution} & \textbf{Maturity} & \textbf{Regulatory anchor} \\
\hline

Technical controls &
RAG, guardrails, uncertainty/abstention, watermarking, logging &
Attach outputs to evidence; mark/contain risk &
RAG \& logging high; guardrails/UQ medium; watermarking emerging &
EU AI Act Arts. 11--12, 15; ISO/IEC 42001 \\

Human oversight &
In-the-loop, on-the-loop, escalation, override &
Place substantive judgement in the path &
Medium (mostly simulated) &
EU AI Act Art. 14; NIST Manage \\

Organisational &
Roles, committees, audit/red team, incident response, management systems &
Embed answerability institutionally &
Medium &
EU AI Act Arts. 9, 17, 27--73; ISO/IEC 42001 \\

Documentation &
Model/system cards, datasheets, evaluation reports, output provenance &
Make decisions investigable post-hoc &
Cards high; output provenance emerging &
EU AI Act Annex IV; NIST Map/Measure \\

\hline
\end{tabular}
}
\end{table*}

\subsection{Choosing Among Mechanism Families for a Given Deployment}

Which mechanisms apply is fixed by three deployment properties, each mapping to a layer. Whether an authoritative reference corpus exists determines Layer 2 grounding the highest-return mechanism where such a corpus exists (clinical guidelines, lending policy, statutory guidance) and unavailable otherwise, shifting the burden to Layer 3. Whether a human sees the output before it is acted upon determines whether Layer 3 review can be load-bearing; where the output acts directly, weight moves to Layer 2 abstention and Layer 4 monitoring. Whether an affected party can be identified determines whether Layer 4 must supply an individual contestation route, which the sectoral instruments of Section 7.4 often mandate, or discharge redress through aggregate monitoring. Two cautions hold throughout: documentation is the cheapest mechanism and the easiest to mistake for accountability, and no single family suffices, since the failure modes of Table~\ref{tab:five_dimensions} are not co-extensive.

\section{Regulatory and Standards Convergence}
\label{sec:rq3}

Between 2022 and 2026, governance shifted from voluntary discourse to binding obligations and certifiable standards  \cite{iso2023iso,lognoul2025regulation}. This section covers the two jurisdictions imposing the most detailed obligations in the corpus, the European Union and the United States, plus the international management-system standard and four sectoral regimes. Other jurisdictions (the United Kingdom's regulator-led approach, China's generative-AI measures) fall outside this mapping. The four sectoral regimes cover the running examples of Section 5.4 plus education, a high-volume deployment setting.

\begin{figure}[!h]
    \centering
    \includegraphics[width=0.8\linewidth]{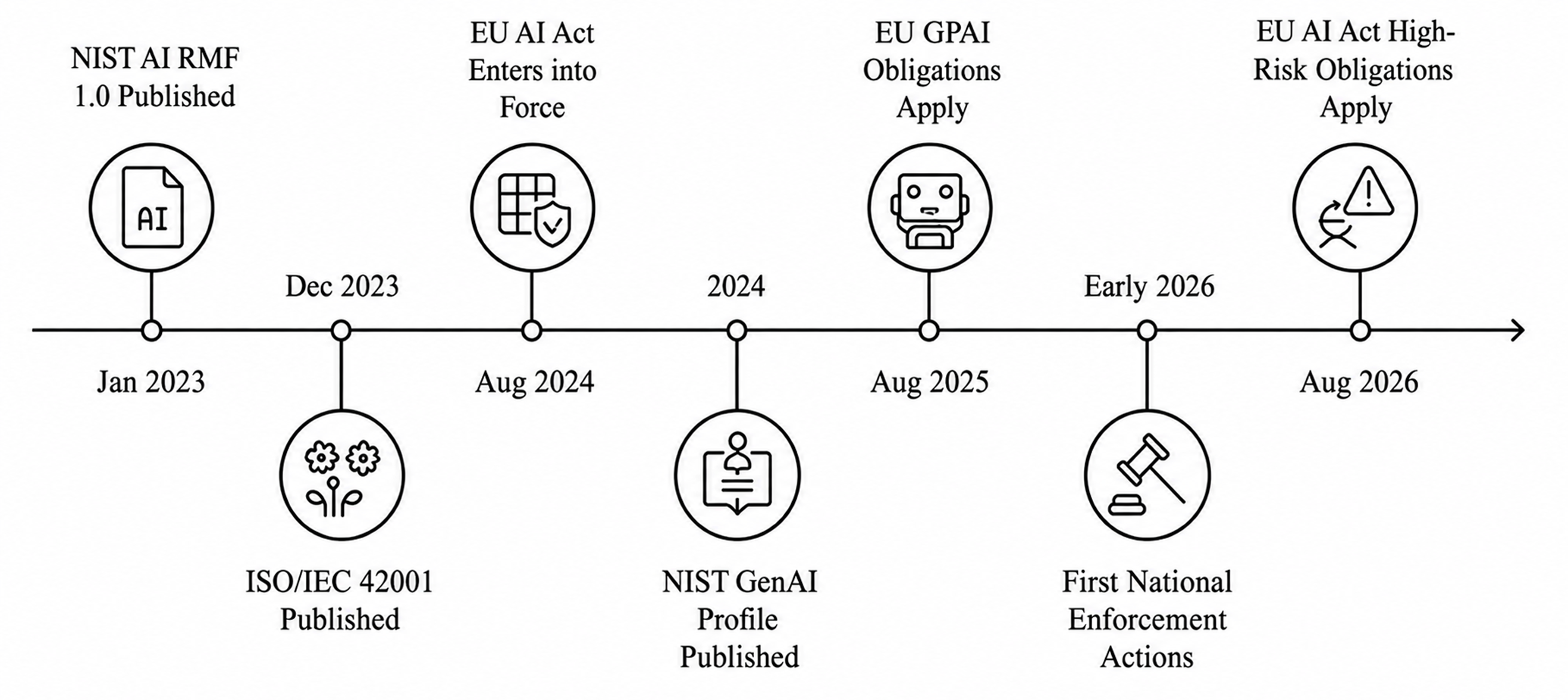}
    \caption{Regulatory and standards landscape timeline. No instrument in scope was published before January 2023, so the timeline begins later than the review window of January 2022 to March 2026.}
    
    \label{fig:F8_LLM}
\end{figure}

Figure \ref{fig:F8_LLM} shows the sequence in which the instruments arrived: the NIST AI RMF (January 2023), ISO/IEC 42001 (December 2023), EU AI Act entry into force (August 2024), the NIST GenAI Profile (2024), EU GPAI obligations (August 2025), and EU AI Act high-risk obligations (August 2026), with sectoral guidance layered along the same period.

\subsection{The EU AI Act}

The EU AI Act is the first comprehensive horizontal AI regulation, classifying systems as prohibited, high-risk, limited, or minimal \cite{lognoul2025regulation, veale2021demystifying}. Chapter V governs General-Purpose AI in two tiers: all providers owe documentation and copyright obligations, while systemic-risk providers owe additional evaluation, adversarial-testing, incident-reporting, and cybersecurity obligations; Layer 1 of LAAF relies on these systemic-risk disclosures as its provenance artefacts.

An LLM in a high-risk context is governed by both regimes: the provider carries the Chapter V obligations and the deployer the high-risk obligations, the regulatory counterpart of the sociotechnical argument of Section 2.5.

Four obligations matter most for the argument of this paper, referred to hereafter as the four core obligations \cite{lognoul2025regulation}. First, oversight: Article 14 requires effective human oversight for high-risk systems \cite{laux2024institutionalised}. Second, documentation and logging: Articles 11 and 12, detailed in Annex IV, require technical documentation and event logging. Third, post-market monitoring: Articles 72 and 73 require post-market monitoring and serious-incident reporting. Fourth, cybersecurity: Article 15 imposes an accuracy, robustness, and cybersecurity obligation.

High-risk obligations under Article 6(2) have applied since 2 August 2026 (safety components under Union harmonisation legislation follow later). Under Article 99, non-compliance carries fines up to €15 million or 3\% of worldwide turnover, with the €35 million or 7\% tier reserved for the prohibited practices of Article 5 \cite{lognoul2025regulation}. A third tier of up to EUR 7.5 million or 1\% of turnover applies to the supply of incorrect, incomplete, or misleading information to authorities, the tier most directly engaged by a documentation and logging framework such as LAAF. For SMEs and start-ups each cap is the lower rather than the higher of the two figures, which bears on the feasibility discussion of Section 11 \cite{smuha2025regulation}.

\subsection{NIST AI RMF and Generative AI Profile}

The NIST AI RMF, published in January 2023, organises practice around four functions: Govern, Map, Measure, and Manage, each subdivided into categories and subcategories of suggested outcomes \cite{ai2023artificial}. The 2024 Generative AI Profile (NIST AI 600-1) is a cross-sectoral companion identifying twelve risks that are unique to or exacerbated by generative AI, among them confabulation, information integrity, information security, data privacy, harmful bias and homogenisation, dangerous, violent, or hateful content, and CBRN information or capabilities \cite{ai2024artificial}. The Profile's operative content is a catalogue of suggested actions, each mapped to an RMF function and subcategory; it is this action-to-function mapping that Section 9.4 uses when aligning LAAF layers to NIST, and the alignment is accordingly at function level rather than at the level of individual controls.

The framework is voluntary but acquired practical force through federal procurement guidance, adoption by sectoral regulators, and reference in state-level AI legislation \cite{united2024advancing,agrawal2026ai}; the RMF itself, as a NIST publication rather than an executive instrument, was unaffected by the executive-branch policy changes of 2025.

\subsection{ISO/IEC 42001}

ISO/IEC 42001, published in December 2023, is the first international standard specifying requirements for an artificial-intelligence management system \cite{iso2023iso}. It adopts the harmonised structure shared by ISO management-system standards, the same structure used by ISO/IEC 27001 for information security, which is why the two integrate readily; 42001 is not derived from 27001 but is a sibling under a common template. Conformity can be certified by third-party bodies operating under national accreditation schemes, on the same basis as other management-system standards.

Its substantive controls, the ISO/IEC 42001 Annex A controls (policy, roles, impact assessment, lifecycle, data, transparency, and third-party relationships), with Annex B guidance — correspond to the mechanisms of Sections 6.1 and 6.4 and are the groupings Section 9.4 intends \cite{mateo2026reference}; adoption evidence remains thin, as accreditation-based certification began only in 2024--2025.

\subsection{Sectoral Regulation}

In healthcare, FDA guidance on AI/ML-enabled device software introduces predetermined change-control plans, and the EMA reflection paper follows the same logic, both interacting with the dynamic-behaviour problem of Section 2.1 \cite{saini2026regulatory,nollen2026artificial}; in consumer finance, CFPB Circular 2023-03 and EBA guidelines require specific, contestable reasons for adverse decisions, aligning with the output-provenance mechanism of Section 6.2 \cite{bureau2022consumer, european2020guidelines}; in education, US Department of Education and UNESCO guidance treats the teaching professional as the accountable party \cite{united2023artificial,holmes2023guidance}; and in the public sector, OECD guidance and EU AI Act Annex~III classify many systems as high-risk \cite{morandin2023recommendation}. Table~\ref{tab:sectoral_guidance} summarises the sectoral guidance across these domains.

\begin{table*}[ht]
\centering
\caption{Sectoral guidance relevant to LLM-based applications. Per sector, the principal guidance, deployment patterns, and the mechanisms each emphasises; the takeaway is that sector calibration is required even under a common horizontal framework.}
\label{tab:sectoral_guidance}
\resizebox{\textwidth}{!}{%
\begin{tabular}{p{2.3cm} p{3.8cm} p{3.5cm} p{5.0cm}}
\hline
\textbf{Sector} & \textbf{Principal guidance} & \textbf{Deployment patterns} & \textbf{Mechanisms emphasised} \\
\hline

Healthcare &
FDA AI/ML SaMD; EMA reflection paper &
Clinical decision support, diagnostics &
Predetermined change control, validation, surveillance \\

Consumer finance &
CFPB Circular 2023-03; EBA &
Credit decisioning, advisory chatbots &
Output traceability, reason-giving, contestability \\

Education &
US Dept. of Education; UNESCO &
Tutoring, assessment, personalisation &
Transparency, instructor oversight, privacy \\

Public sector &
OECD; EU AI Act Annex III &
Service delivery, benefits, hiring &
Risk classification, oversight, contestability \\

\hline
\end{tabular}
}
\end{table*}

\subsection{Comparative Analysis Across Frameworks}

Table~\ref{tab:framework_comparison} compares the four instrument types ("types" rather than "families", which Section 6 reserves for mechanisms). On the authors' reading, they are complementary: the EU AI Act supplies binding force, the NIST AI RMF operational specificity, ISO/IEC 42001 certifiable conformity, and sectoral guidance domain calibration; no one instrument supplies all four.

All four converge on the lifecycle character of accountability, requiring that obligations attach before, during, and after deployment rather than at a single approval point. The literature lags the instruments on operational specifics for the four core obligations, which Section 8.3 reports as a finding.

Because the four types agree on the lifecycle requirement and differ mainly in force and specificity, a single architecture assigning artefacts to lifecycle positions can satisfy all four simultaneously, the claim Section 9.4 makes concrete layer by layer.

\begin{table*}[!h]
\centering
\caption{Comparative analysis of regulatory and standards instrument types. Aligns the four instrument types across legal status, scope, and the four core obligations of Section 7.1; the takeaway is convergence on common requirements, which in principle allows one architecture to serve several jurisdictions, though the design and audit cost of doing so has not been measured empirically.}
\label{tab:framework_comparison}
\begin{tabular}{p{2.8cm} p{2.5cm} p{2.5cm} p{2.5cm} p{2.5cm}}
\hline
\textbf{Dimension} & \textbf{EU AI Act} & \textbf{NIST AI RMF} & \textbf{ISO/IEC 42001} & \textbf{Sectoral} \\
\hline

Legal status &
Binding, penalties &
Voluntary, influential via procurement and state reference &
Voluntary, certifiable &
Mixed: binding where statutory, advisory otherwise \\

Scope &
Horizontal &
Horizontal &
Horizontal &
Sector-specific \\

Human oversight &
Mandatory (Art. 14) &
Govern and Manage; GenAI Profile actions &
Operational controls &
Mandatory in regulated sectors \\

Documentation/logging &
Mandatory (Arts. 11--12, Annex IV) &
Map/Measure &
Documented information &
Sector-specific records \\

Post-deployment monitoring &
Mandatory (Arts. 72--73) &
Measure and Manage &
Performance evaluation &
Post-market surveillance \\

Cybersecurity &
Mandatory (Art. 15) &
Map, Measure, Manage; GenAI Profile actions &
Annex controls &
Sector-specific \\

\hline
\end{tabular}
\end{table*}

\section{Consolidated Findings Against the Review Questions}
\label{sec:findings}

This section consolidates the answers to the four review questions against the coded corpus, then reports what persists once the three bodies of evidence are read together: four gaps in what the literature specifies, and five tensions it cannot resolve.% label added by Pierre

\subsection{Conceptual Findings}
\label{sec:findrq1} %label added by PD

The corpus converges on an answerability-oriented conception: among studies that
define accountability explicitly rather than invoking it, the actor forum
relation of Section 5.1 is the dominant framing \cite{bovens2007analysing,
novelli2024accountability}. We report this qualitatively; the extraction template
did not code definitional stance as a separate countable field, which Section 11
records as a limitation.

Five strands contribute to this convergence, and the grouping is our synthesis rather than an established typology. The answerability strand supplies the actor–forum relation \cite{novelli2024accountability}. The constituted-system strand relocates the object of governance from the model to the assembled sociotechnical arrangement (Section 2.5) \cite{nabben2024ai,weidinger2023sociotechnical}. The attributability-gap strand identifies how confident model output erodes substantive human ownership of decisions (Section 5.2) \cite{zeiser2024owning}. The dual-phase generative strand shows how the human actor recedes behind a responsive interface (Section 5.2) \cite{elliott2025evolving}. The risk-governance strand supplies the organisational apparatus through which answerability is institutionalised (Section 5.4) \cite{schuett2025three}.

The answer to RQ1 is that accountability is an answerability relation between an actor and a forum, judged against standards through a defined process with consequences, resolved into the five dimensions of Section 5.3 and anchored in sociotechnical thinking. Two conceptual gaps remain: the corpus converges on what accountability is without converging on how its dimensions could be measured, and it treats enforcement-and-redress least often; both are developed in Section 8.5.

\subsection{Mechanism Findings}
Any accountable application requires mechanisms at all
four layers of Section 2.6 operating together, and maturity declines as one moves from the model toward the forum. Detection is well formalised (Equations (1)--(4)) and grounding, retrieval, and logging are production-standard. Guardrails and uncertainty quantification are of medium maturity, and watermarking and output-level provenance remain emerging, with no standardised scheme or schema in production use. Maturity is thus highest where mechanisms sit closest to the model and lowest exactly where the regulatory obligations of Section 7 are most specific.

\subsection{Regulatory Findings}
\label{sec:findrq3} %label added by PD

There is a convergence towards common lifecycle requirements across the EU AI Act, the NIST AI RMF and its GenAI Profile, ISO/IEC 42001, and sectoral guidance \cite{lognoul2025regulation,ai2023artificial,iso2023iso}. In principle this allows one architecture to serve several jurisdictions, which the literature treats as a reduction in the cost of designing for multiple regimes \cite{roberts2024global,smuha2021race}; that cost has not been measured
empirically, and no study in the corpus attempts it. 
 
The literature is still lacking in operational detail, including: output-provenance templates; substantive oversight criteria under Article~14; and the contestability of adverse decisions.

\subsection{Cross-Cutting Patterns}

Reading the three bodies of evidence together yields three patterns that no single review question isolates.

First, mechanism maturity declines monotonically with distance from the model, while regulatory specificity increases with it. Detection is formally characterised and grounding, retrieval, and logging are production-standard, all at Layers 1 and 2; oversight and governance mechanisms at Layers 3 and 4 are specified but evaluated mostly under simulated conditions. The regulatory obligations run the other way, being most prescriptive precisely at Article 14 oversight and Articles 72 and 73 post-market monitoring. The two curves cross, and the crossing point is where deployment risk concentrates: obligations are most specific exactly where the evidence base is thinnest.

Second, the four instrument types converge on the lifecycle character of accountability and diverge only on legal force and operational specificity, which is what makes a single architecture serving several jurisdictions feasible in principle, though the design and audit cost of doing so is not measured in any study in the corpus.

Third, evaluation conditions are systematically favourable across every mechanism family. Retrieval is assessed on curated corpora, guardrails against static attack libraries, uncertainty methods without reviewer-interpretability testing, and oversight under simulation. Adversarial, degraded, and longitudinal conditions are near-absent, which bounds what any maturity claim in Table~\ref{tab:mechanism_families} can support.
\subsection{Four Persistent Gaps}
\label{sec:gaps} %label added by PD

Four gaps persist across the corpus; Table~\ref{tab:persistent_gaps} converts each into a design requirement. First, human oversight is under-specified: the human in the loop is invoked without specifying what is reviewed, when, with what information, and with what authority \cite{green2022flaws, sterz2024quest}. Second, shared metrics for the accountability dimensions of Section 5.3 are absent \cite{novelli2024accountability,yehudai2025survey}. Third, the technical and governance literatures remain disconnected \cite{mokander2024auditing,hacker2023regulating}. Fourth, empirical evaluation is limited: few studies test whether oversight intercepts harmful outputs or incident response produces durable change \cite{yehudai2025survey,raji2020closing}.

\begin{table*}[!h]
\centering
\caption{Persistent gaps and resulting requirements for LAAF. Each gap is documented to specific sections and converted into a concrete requirement imposed on the framework; the takeaway is that the gaps are specific and tractable, not diffuse.}
\label{tab:persistent_gaps}
\resizebox{\textwidth}{!}{%
\begin{tabular}{p{4.0cm} p{2.5cm} p{7.0cm}}
\hline
\textbf{Persistent gap} & \textbf{Where documented} & \textbf{Requirement imposed on LAAF} \\
\hline

Under-specification of human oversight &
1.2, 7.1, 8.5

&

Specify what, when, with what information, with what authority \\

Absence of shared accountability metrics &
7.5, 8.5 

&
Define layer-specific accountability indicators \\

Disciplinary disconnection &
1.2, 6, 7 &
Integrate mechanisms from the technical and governance literatures 

\\

Limited empirical evaluation &
Sections 6--8 &
Articulate evaluation criteria and validation pathways \\

\hline
\end{tabular}
}
\end{table*}

\subsection{Five Structural Tensions}
\label{sec:tensions} %label added by PD

The favourable evaluation conditions reported in Section 8.4 are why Section 10 specifies evaluation criteria rather than claiming validation. The mechanisms carry limitations of their own.

RAG is evaluated on well-curated corpora unrepresentative of messy real retrieval \cite{gao2023retrieval,mishra2024fine}. Guardrails are tested against static attack libraries and fail against adaptive adversaries \cite{yang2024assessing, greshake2023not}. Uncertainty methods are rarely tested for interpretability to non-technical reviewers \cite{lin2022teaching, kuhn2023semantic}. Oversight studies rely on simulated conditions, override rates rarely being measured in deployment \cite{bansal2021does, inkpen2023advancing}. Output-level provenance remains weakest, with no standardised template \cite{wang2026agent}. 

Beyond both sit five structural tensions (the authors' synthesis): transparency conflicts with competitive and security interests, since accountability-enabling disclosures also enable copying and attack; automation with substantive oversight, since review erodes the efficiency justifying deployment; standardisation with contextual fit, since a common template underspecifies each sector; traceability with privacy, since the provenance record is itself sensitive data; and deployment speed with governance maturity, since capability arrives before its governing apparatus. Table~\ref{tab:structural_tensions} summarises each with LAAF's design response.

Taken together, the four persistent gaps of Section 8.5 and the five tensions above constitute the design brief for the framework: the gaps state what it must supply, and the tensions state what it must navigate rather than resolve.

\begin{table*}[!h]
\centering
\caption{Structural tensions and how LAAF addresses them. The five trade-offs any deployment must navigate, with the LAAF design response; the takeaway is that accountability is a navigation problem, not a checklist.}
\label{tab:structural_tensions}
\resizebox{\textwidth}{!}{%
\begin{tabular}{p{3.5cm} p{3.5cm} p{6.0cm}}
\hline
\textbf{Tension} & \textbf{Where most visible} & \textbf{How LAAF addresses it} \\
\hline

Transparency vs.\ competitive/security &
Documentation, watermarking, security &
Separate internal/external disclosure surfaces in Layer~4 \\

Automation vs.\ substantive oversight &
Oversight, decision support &
Distinguish review regimes; require explicit justification \\

Standardisation vs.\ contextual fit &
Audit, standards, sectoral &
Sector-agnostic architecture with sector specialisations \\

Traceability vs.\ privacy &
Provenance, logging, privacy &
Governed provenance and audit assets in Layer~4 \\

Deployment speed vs.\ governance maturity &
Capability vs.\ risk governance &
Incremental adoption by layer, with the accountability properties forgone at each stage stated explicitly \\

\hline
\end{tabular}
}
\end{table*}

\section{Synthesis: Toward an Integrated Accountability Architecture}
\label{sec:laaf} %label added vy PD

This section consolidates the findings of Section 8 into a single account. To make the layers concrete, consider a clinical query traced end to end through the four layers defined in Section 9.2:

a clinician submits summarise this patient's anticoagulation options. Layer 1 fixes the model, version, and systemic-risk disclosures; Layer 2 grounds the answer in retrieved guidelines with per-claim citations and an uncertainty score, abstaining on a low-confidence contraindication; Layer 3 places a named clinician in the path, who overrides one unsupported sentence and records the reason; Layer 4 logs the override, and a later audit of a recurring abstention pattern triggers a documented configuration change and, where a patient was affected, a redress pathway. One input thus produces a traceable chain from provenance to redress.

\subsection{Principles Emerging from the Literature}\label{sec:principles} %label added by PD

LAAF specifies the layers, responsibilities, information flows, and
intervention points any accountable deployment should make explicit, leaving
the software stack, vendor, and jurisdiction open. Built against the requirements of Section 8.5 and the tensions of Section 8.6, it rests on five principles: layered (no single mechanism captures the sociotechnical character), lifecycle-oriented (behaviour shifts with updates and context), distributed across named actors (diffusion of responsibility is a central failure mode), regulator-mappable (obligations are now binding), and navigable across tensions. One scope condition applies throughout. LAAF assumes a single output reviewed by a human before it is acted upon, which is the decision-support setting of the three running examples of Section 5.4. Agentic and tool-calling systems break this assumption: an action may complete before any review point exists, so provenance must attach to actions rather than to text, Layer 3 becomes pre-authorisation and action budgets rather than output review, and Layer 4 requires rollback and interruption capability. Extending LAAF to that setting is a priority in Section 10. 

\subsection{Architecture Overview}\label{sec:architecture} %label added by PD

\begin{figure}[!h]
    \centering
    \includegraphics[width=0.9\linewidth]{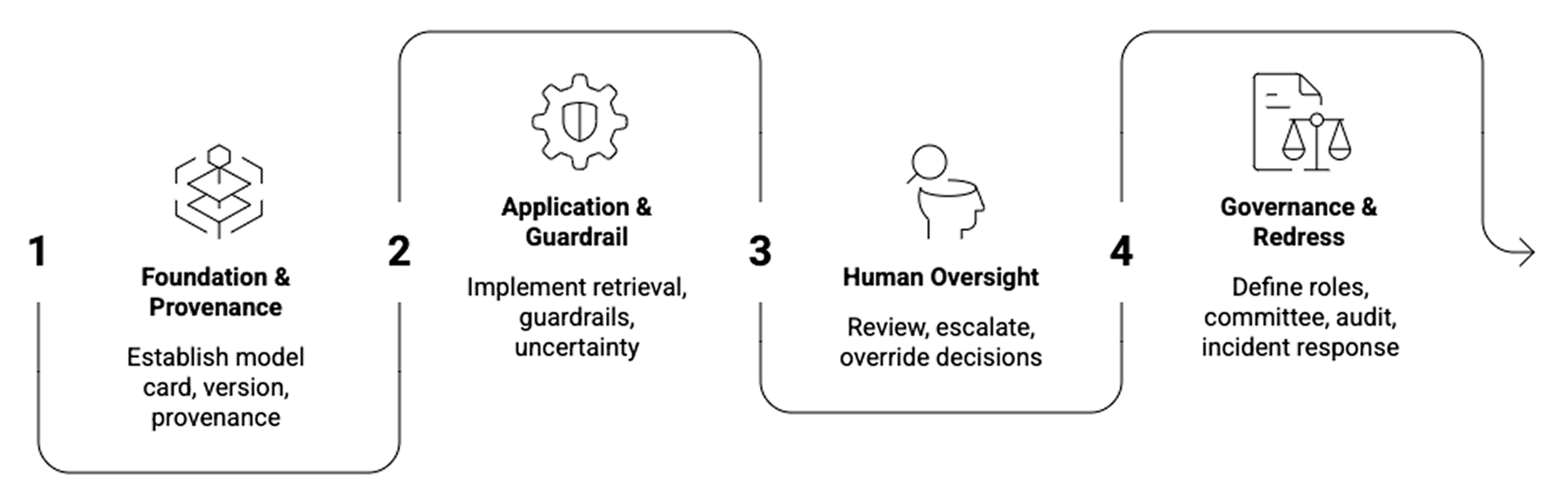}
    \caption{The Layered Accountability Architecture Framework (LAAF): four layers spanned by three cross-cutting properties (traceability, role clarity, continuous monitoring), ordered by distance from the model. Evidence flows upward, constraining the layer above; binding the layer below. }
    \label{fig:F9_LLM}
\end{figure}

LAAF is organised as a vertical stack of four layers (Figure~\ref{fig:F9_LLM}), namely Foundation Model and Provenance, Application Logic and Guardrails, Human Oversight and Review, and Governance, Audit, and Redress, cross-cut by traceability, role clarity, and continuous monitoring.

The ordering is by distance from the model, terminating at Layer 4 where accountability resolves once a harm has occurred.

Accountability runs in both directions between adjacent layers: each layer's artefacts constrain the configuration of the layer above as evidence, while policy set above binds the review and guardrail regimes below. Authority flows downward only within the deploying organisation's span of control, from Layer 4 to Layer 2. At the Layer 2 to Layer 1 boundary it cannot be exercised at all, because the foundation-model provider lies outside the deployer's authority, and recourse there is limited to model selection, contractual terms, and the obligations regulation places on the provider. That boundary is where accountability most often breaks in practice. Table~\ref{tab:laaf_layers} summarises the layers.

\begin{table*}[!h]
\centering
\caption{The four layers of LAAF: accountability function, principal artefacts, and named accountable actor per layer.}
\label{tab:laaf_layers}
\resizebox{\textwidth}{!}{%
\begin{tabular}{p{2.2cm} p{3.5cm} p{4.5cm} p{4.0cm}}
\hline
\textbf{Layer} & \textbf{Accountability function} & \textbf{Principal artefacts} & \textbf{Named accountable actor} \\
\hline

L1 Foundation Model \& Provenance &
Establish verifiable identity of model and training basis &
Model card, training-data record, version, systemic-risk disclosures &
Foundation-model provider; fine-tuning party \\

L2 Application Logic \& Guardrails &
Ground generation, enforce policy, mark uncertainty, log behaviour &
Retrieval pipeline, guardrails, uncertainty scores, abstention rules, output provenance &
Application developer; deploying organisation \\

L3 Human Oversight \& Review &
Place substantive human judgement in the path &
Review workflow, escalation pathway, override authority, regime selector &
Deploying organisation; domain professional \\

L4 Governance, Audit \& Redress &
Allocate roles, govern, audit, respond to incidents &
Role allocation, governance committee, audit, incident response, management system &
Deploying organisation (corporate entity) \\

\hline
\end{tabular}
}
\end{table*}

\subsection{The Three Cross-Cutting Properties}
\label{sec:properties} %label added by PD

The three properties are conditions each layer must satisfy, corresponding to what accountability requires simultaneously: evidence that a decision occurred as recorded, a person answerable for it, and vigilance that the arrangement holds. Contestability is a Layer 4 artefact rather than a fourth property, being exercised externally rather than maintained by every layer. The three derive from the regulatory convergence of Section 7.5: traceability from Annex IV and Article 12, role clarity from Articles 16 and 26, and continuous monitoring from Articles 72 and 73.

Traceability is the property that every consequential decision, automated or human, can be reconstructed after the fact. Role clarity resolves each override, configuration change, or sign-off to a named individual, going beyond the Act's provider–deployer distinction.

Continuous monitoring is the property that behaviour is observed beyond deployment and deviations trigger defined responses: pre-defined thresholds on abstention, override, retrieval-failure, and drift rates; automatic escalation to Layer 3 on breach; and an incident record at Layer 4 when escalation does not resolve it. 

Table~\ref{tab:cross_cutting_properties} specifies how each property is realised per layer. One asymmetry deserves note: Layer 1 monitoring depends on provider notifications the deployer typically cannot compel; recourse is contractual (notification clauses) and regulatory (Chapter V transparency), a known limitation Section 10 flags as a governance priority.

\begin{table*}[!h]
\centering
\caption{Cross-cutting properties and their realisation per layer. Shows the concrete artefact realising each property in each layer; the takeaway is that the properties stitch the four layers into a single auditable chain.}
\label{tab:cross_cutting_properties}
\resizebox{\textwidth}{!}{%
\begin{tabular}{p{2.3cm} p{3.0cm} p{3.2cm} p{3.0cm} p{3.2cm}}
\hline
\textbf{Property} & \textbf{Layer 1} & \textbf{Layer 2} & \textbf{Layer 3} & \textbf{Layer 4} \\
\hline

Traceability &
Model/version ID, training-data record &
Output provenance, retrieval citations, uncertainty &
Review record, override record &
Audit record, incident record, minutes \\

Role clarity &
Provider \& fine-tuner named &
Developer \& deployer named &
Reviewer \& override authority named &
Named individuals per governance role \\

Continuous monitoring &
Update notifications, risk reassessment &
Logging hooks, drift detection &
Review/override pattern analysis &
Incident cycles, audit cycles, improvement loop \\

\hline
\end{tabular}
}
\end{table*}

\subsection{Regulatory Integration}
\label{sec:regmapping} %label added by PD

LAAF's fourth design principle is regulator-mappability. Layer~1 aligns with EU AI Act Chapter~V, the NIST Map function, and ISO/IEC 42001 design-and-development controls \cite{ai2023artificial,iso2023iso}. Layer~2 aligns with EU AI Act Articles~11--12, NIST Map and Measure, and ISO/IEC 42001 operational controls \cite{lognoul2025regulation,ai2024artificial}. Layer~3 aligns with EU AI Act Article~14, NIST Manage, and ISO/IEC 42001 roles-and-competence controls \cite{laux2024institutionalised}. Layer~4 aligns with EU AI Act Articles~9, 17, 72, and 73, all four NIST functions, and the full ISO/IEC 42001 clause structure \cite{mateo2026reference}. Table~\ref{tab:laaf_regulatory_mapping} presents the mapping with cybersecurity anchors and layer-specific indicators.

\begin{table*}[!h]
\centering
\caption{Mapping of LAAF layers to regulatory and cybersecurity requirements, with indicators. Per layer, the EU/NIST/ISO anchors, the OWASP LLM Top 10 (2025) categories, and a measurable indicator; the takeaway is that LAAF is regulator-mappable and security-integrated at every layer, with concrete metrics.}
\label{tab:laaf_regulatory_mapping}
\resizebox{\textwidth}{!}{%
\begin{tabular}{p{0.8cm} p{1.2cm} p{1.0cm} p{2.0cm} p{4.8cm} p{2.8cm}}
\hline
\textbf{Layer} & \textbf{EU AI Act} & \textbf{NIST} & \textbf{ISO/IEC 42001} & \textbf{OWASP LLM Top 10 (2025)} & \textbf{Layer indicator} \\
\hline

L1 &
Ch. V, Annex IV &
Map &
Design/development controls &
LLM03 Supply Chain; LLM04 Data \& Model Poisoning &
Completeness of model/training disclosures \\

L2 &
Arts. 11--12, Annex IV &
Map, Measure &
Operational controls &
LLM01 Prompt Injection; LLM02 Sensitive Info Disclosure; LLM05 Improper Output Handling; LLM06 Excessive Agency; LLM08 Vector \& Embedding Weaknesses \& LLM09 Misinformation &
Proportion of outputs with citations \& uncertainty \\

L3 &
Art. 14 &
Manage &
Roles \& competence &
LLM07 System Prompt Leakage &

Override rate, review time, regime justification \\

L4 &
Arts. 9, 17, 72--73 &
Govern (all) &
Leadership, planning, evaluation, improvement &
LLM10 Unbounded Consumption (org level); all categories &
Incident closure rate, audit-cycle completion \\

\hline
\end{tabular}
}
\end{table*}

\subsection{Cybersecurity Integration}
\label{sec:cybersecurity} %label added by PD

A system whose outputs cannot be defended against manipulation cannot answer
for what it produces, since an output an adversary may have induced is one for
which no actor can account. We advance this as our own argument rather than as
a finding of the corpus, where the two literatures remain largely separate
\cite{das2025security}. Cybersecurity is therefore treated as integral to accountability rather than adjacent to it, and each layer is aligned to the relevant categories of the OWASP LLM Top 10 (2025) \cite{huang2024genai}, with the general link between LLM security and trustworthy deployment supported by the surveyed literature \cite{da2024survey}.

The per-layer mapping to OWASP categories is recorded in Table~\ref{tab:laaf_regulatory_mapping}. Layer 2 carries the largest share, since retrieved content is itself an injection surface, and Layer 3 additionally requires reviewers trained to recognise adversarial outputs such as instruction-like text in retrieved passages and citations resolving to no source. Governance at Layer 4 institutionalises response by integrating the ISO/IEC 27001 information-security management system with the ISO/IEC 42001 AI management system, giving unified incident response, tamper-resistant logging, and monitoring for unbounded consumption. The three cross-cutting properties carry security implications of their own: traceability supports forensic reconstruction, role clarity assigns response ownership, and continuous monitoring detects both drift and indicators of compromise.

\subsection{Sector-Specific Instantiation of the Architecture}

Table~\ref{tab:laaf_sector_cases} shows how each layer specialises across the three running examples of Section 5.4 while the four-layer structure stays constant. Sector-specific instantiation beyond these three is listed as a priority in Section 10.

\begin{table*}[!h]
\centering
\caption{LAAF sector specialisations across three case studies. Shows how each layer specialises across healthcare, finance, and public sector while the four-layer/three-property structure stays constant; the takeaway is sector-agnostic structure with sector-calibrated content.}
\label{tab:laaf_sector_cases}
\resizebox{\textwidth}{!}{%
\begin{tabular}{p{2.3cm} p{4.0cm} p{4.0cm} p{4.0cm}}
\hline
\textbf{LAAF element} & \textbf{Healthcare} & \textbf{Finance} & \textbf{Public sector} \\
\hline

L1 priorities &
Clinical safety in training data &
Fair lending in training data &
Democratic neutrality in training \\

L2 guardrails &
Clinical accuracy, citation grounding &
Discriminatory features, data leakage &
Political balance, source verification \\

L3 reviewer &
Clinician with medical authority &
Credit officer with adverse-action knowledge &
Civil servant with verification duty \\

L4 governance &
Clinical-AI lead + hospital quality &
Credit risk + fair lending + consumer protection &
Digital service + DPO + accountability officer \\

Regulatory anchors &
EU AI Act high-risk, FDA, EMA &
CFPB, EBA, EU AI Act Annex III &
OECD, EU AI Act Annex III \\

Sector-specific risk &
Diagnostic error, patient harm &
Discriminatory denial, financial harm &
Citizen rights, democratic harm \\

\hline
\end{tabular}
}
\end{table*}

\subsection{LAAF versus Existing Frameworks}
\label{sec:frameworkcomparison}

Table~\ref{tab:laaf_comparison} compares the architecture against fourteen existing governance instruments on seven criteria drawn from Section 8: layered decomposition and sector applicability from the disciplinary-disconnection gap and the standardisation-versus-context tension, human oversight from the oversight under-specification gap, regulatory mappability from Section 7.5, implementation roadmap from the deployment-speed tension, evaluation criteria from the empirical-evaluation gap, and cybersecurity from the transparency-versus-security tension. Coverage concentrates at oversight and regulatory mappability; cybersecurity is addressed as a design element by one instrument only, and sector calibration by three. To our knowledge no existing instrument addresses all seven together.

\begin{table*}[!h]
\centering
\caption{Comparative analysis of LAAF against existing governance frameworks. Each framework is scored on the seven criteria listed at the head of this subsection, where \textbf{$\checkmark$} denotes the criterion is addressed as a design element, P that it is discussed without being operationalised, and × that it is absent. To our knowledge, LAAF is the only entry covering all seven. Scores are author-assigned against this rubric without a second independent rater, which Section 11 records as a limitation. The Eval. column records whether evaluation criteria are articulated, not whether the framework has been empirically validated; no entry in the table, including LAAF, reports empirical validation in deployment. Layered denotes decomposition of accountability into ordered strata each with a distinct accountable actor, rather than into process phases or clause groups.}

\label{tab:laaf_comparison}
\resizebox{\textwidth}{!}{%
\begin{tabular}{p{0.8cm} p{3.2cm} c c c c c c c}
\hline
\textbf{Ref.} & \textbf{Framework} & \textbf{Layered} & \textbf{Oversight} & \textbf{Cyber} & \textbf{Reg. map} & \textbf{Sector} & \textbf{Roadmap} & \textbf{Eval. criteria} \\
\hline

{\cite{mokander2024auditing}} & Three-layered LLM audit

& $\checkmark$ & P & $\times$ & P & $\times$ & $\times$ & $\times$ \\

{\cite{ai2023artificial}} & NIST AI RMF
& $\times$ & $\checkmark$ & P & P & $\times$ & $\times$ & P \\

{\cite{ai2024artificial}} & NIST GenAI Profile
& $\times$ & $\checkmark$ & $\checkmark$ & P & P & $\times$ & $\times$ \\

{\cite{iso2023iso}} & ISO/IEC 42001
& $\checkmark$ & $\checkmark$ & P & P & $\times$ & $\times$ & P \\

{\cite{bommasani2023foundation}} & Transparency Index
& $\times$ & $\times$ & $\times$ & P & $\times$ & $\times$ & $\times$ \\

{\cite{schuett2025three}} & Risk governance
& P & P & $\times$ & $\times$ & $\times$ & $\times$ & $\times$ \\

{\cite{jongepier2022explanation}} & Reasoned-explanation right
& $\times$ & $\checkmark$ & $\times$ & P & P & $\times$ & $\times$ \\

{\cite{raji2020closing}} & Closing accountability gap
& P & $\times$ & $\times$ & $\times$ & $\times$ & $\times$ & $\times$ \\

{\cite{floridi2022capai}} & capAI conformity assessment
& $\checkmark$ & $\checkmark$ & P & P & P & $\checkmark$ & P \\

{\cite{laux2024institutionalised}} & Article 14 operationalisation
& $\times$ & $\checkmark$ & $\times$ & P & P & $\times$ & $\times$ \\

{\cite{morandin2023recommendation}} & OECD AI Principles
& $\times$ & P & $\times$ & $\times$ & $\times$ & $\times$ & $\times$ \\

{\cite{leon2026lifecycle}} & AI lifecycle accountability
& $\checkmark$ & P & $\times$ & P & $\times$ & P & P \\

\cite{mokander2023operationalising} & Ethics-based auditing
& P & P & $\times$ & P & $\times$ & $\times$ & P \\

\cite{naser2025auditing} & Training-data auditing
& P & $\checkmark$ & $\times$ & P & $\times$ & $\times$ & $\times$ \\

\textbf{Ours} & \textbf{LAAF}
& \textbf{$\checkmark$}
& \textbf{$\checkmark$}
& \textbf{$\checkmark$}
& \textbf{$\checkmark$}
& \textbf{$\checkmark$}
& \textbf{$\checkmark$}
& \textbf{$\checkmark$} \\

\hline
\end{tabular}
}
\end{table*}

\section{Open Questions and Research Priorities}
\label{sec:agenda} %label added by PD

The review leaves four kinds of question open, and they are ordered here by what would have to change for each to be answered: new measurement instruments, new technical artefacts, new governance practice, and new collaboration across communities. Each follows directly from a gap in Section 8.5.

\subsection{Measurement} Accountability benchmarks that evaluate oversight substantiveness, governance closure, and enforcement responsiveness rather than only narrow performance \cite{raji2020closing}; operational metrics for the five dimensions of Section 5.3 \cite{novelli2024accountability}; and empirical case-study research on accountability practice in real deployments \cite{yehudai2025survey}.
Evaluating the architecture of Section 9 requires four criteria for which no instrument currently exists: completeness, whether all layers and properties are operational; traceability, whether a decision can be reconstructed end to end; oversight substantiveness, whether review meaningfully alters outputs; and governance closure, whether incidents lead to recorded systemic change. A cheaper first step is retrospective, asking of each documented deployment failure which layer would have intercepted it and which artefact would make it reconstructible afterwards, though retrospective attribution from published accounts cannot establish what a layer would have caught.

\subsection{Technical artefacts} Auditable RAG producing a structured record linking each claim to retrieved passages with explicit faithfulness scores \cite{liang2024empowering,mishra2024fine}; calibrated, human-meaningful uncertainty aligned with reviewer decision categories \cite{lin2022teaching,kuhn2023semantic}; and standardised output-level provenance metadata making Layer 2 interoperable \cite{wang2026agent}.

\subsection{Governance and policy} Operationalising effective human oversight under Article 14 \cite{laux2024institutionalised}; criteria for assigning outputs to review regimes, so that a forum can ask not only whether an output was reviewed but whether it should have been, and so that projected review volume is matched against stated reviewer capacity; sector-specific instantiation of Layers 3 and 4, in particular calibration of review regimes and escalation thresholds to sectoral duty-of-care standards in healthcare, consumer finance, and public administration; redress mechanisms satisfying the enforcement-and-redress dimension \cite{jongepier2022explanation,raji2020closing}; integrating LAAF with ISO/IEC 42001 certification \cite{iso2023iso,mateo2026reference}. and extending Layers 2 to 4 to agentic and tool-calling deployments \cite{van2026agentic,wang2026agent}.

\subsection{Interdisciplinary and cybersecurity} Shared vocabulary across communities \cite{mokander2024auditing}; joint computer-science and human-factors work on Layer 3 design \cite{bansal2021does,inkpen2023advancing}; sustained academic–regulator engagement \cite{raji2020closing,hacker2023regulating}; adversarial-robustness benchmarks for accountability mechanisms \cite{yang2024assessing,greshake2023not}; and integrated AI-and-information-security management combining ISO/IEC 42001 and 27001 \cite{ali2026operationalising,iso2023iso}.

\section{Limitations and Threats to Validity}

Following established practice, threats are organised into four categories \cite{arvanitou2021software,xiao2019guidance}. Construct threats include the imprecision of the term accountability and the boundary of the system under governance, mitigated by separating conceptual from mechanism extraction \cite{novelli2024accountability,edenberg2023disambiguating}, and that definitional stance was not coded as a countable field. Internal threats include researcher and publication bias, mitigated by prespecified criteria, application of the rubric in two independent passes separated in time, and direct retrieval of regulatory documents
\cite{sohrabi2021prisma}.

A residual internal threat is that the framework comparison of Table~\ref{tab:laaf_comparison} is author-scored, mitigated by publishing the rubric but not eliminated. External threats include the temporal scope of January 2022 to March 2026, which
excludes work appearing between the search cutoff and publication; the
conjunction of a sector term with the accountability block in the search string,
which excludes general accountability work not framed in a named sector; the
English-only constraint; the non-registration of the protocol (Section 3.1); and the treatment of preprints as a separately documented stratum rather than through systematic screening (Section 3.3). LAAF is most
directly applicable to deployments under the frameworks analysed in Section 7.

Conclusion threats arise from the qualitative nature of the synthesis, mitigated by source-appropriate quality assessment, and the residual limitation is that quantitative comparison of accountability mechanisms across studies is not yet possible, which is itself one of the review's findings \cite{yehudai2025survey,raji2020closing}. The framework carries threats distinct from those of the review. LAAF would be disconfirmed by a deployment that satisfies all four layers and all three properties and is nonetheless unaccountable in a way the architecture cannot diagnose, for example a deployment in which every artefact is present and the contested decision rule itself is the object of the objection. LAAF is therefore falsifiable, and the class of cases that falsify it is the class in which accountability fails for substantive rather than procedural reasons. Its scope conditions are equally specific: it does not cover agentic deployments (Section 9.1), it assumes governance capacity that small deployers may lack, and it is mapped only to the instruments of Section 7.

\section{Conclusion}
\label{sec:conclusion}

The question with which this paper began, when an LLM output contributes to harm, who is answerable, has a single sustained answer. The paper established that LLMs are sociotechnical systems whose harms are distributed across the lifecycle and whose hallucinations are structural rather than incidental; synthesised mechanisms across four families; mapped the literature onto the binding regulatory landscape; and consolidated four persistent gaps and five structural tensions into LAAF, a regulator-mappable, sector-agnostic, four-layer architecture with three cross-cutting properties, integrated cybersecurity aligned with the OWASP LLM Top 10 (2025), and a comparison against fourteen frameworks. The contribution is that accountability for LLM applications is specifiable and decomposable, reducible neither to a single mechanism nor to a one-time decision, and that the resulting gaps are specific enough to be converted into an architecture. The review does not establish that the architecture works. Section 10 sets out
the retrospective checks available and the deployment evaluation still required,
and no claim of effectiveness is made here.

\section*{Data Availability}

The search strategy, inclusion criteria, and quality-assessment rubric are reported in full in
Section 3, and the scoring rubric for Tables~\ref{tab:related_work} and \ref{tab:laaf_comparison} is given in Section 3.7. A replication package
containing the record of the 122 included studies, the completed extraction sheet, the per-study
quality scores, and the comparison rubrics with their scored cells accompanies this submission.

\section*{AI Declaration}
Use of generative AI: Generative AI tools were used for language editing of author-drafted text. They were not used to identify, screen, or select studies, to extract or code data, to assign quality or comparison scores, or to generate the review's findings.

\section*{Author Contributions}

All authors contributed equally to the conception, analysis, interpretation, and writing of this manuscript.
%All authors contributed to the conception and design of the study; contributed to the technical review, analysis, and interpretation of results; participated in drafting and critically revising the manuscript for important intellectual content.

\bibliographystyle{plain}
\bibliography{reference}

@article{constantinescu2021understanding,
  title={Understanding responsibility in Responsible AI. Dianoetic virtues and the hard problem of context},
  author={Constantinescu, Mihaela and Voinea, Cristina and Uszkai, Radu and Vic{\u{a}}, Constantin},
  journal={Ethics and Information Technology},
  volume={23},
  number={4},
  pages={803--814},
  year={2021},
  publisher={Springer}
}

@article{li2024chatgpt,
  title={ChatGPT in healthcare: a taxonomy and systematic review},
  author={Li, Jianning and Dada, Amin and Puladi, Behrus and Kleesiek, Jens and Egger, Jan},
  journal={Computer Methods and Programs in Biomedicine},
  volume={245},
  pages={108013},
  year={2024},
  publisher={Elsevier}
}

@article{omar2025sociodemographic,
  title={Sociodemographic biases in medical decision making by large language models},
  author={Omar, Mahmud and Soffer, Shelly and Agbareia, Reem and Bragazzi, Nicola Luigi and Apakama, Donald U and Horowitz, Carol R and Charney, Alexander W and Freeman, Robert and Kummer, Benjamin and Glicksberg, Benjamin S and others},
  journal={Nature Medicine},
  volume={31},
  number={6},
  pages={1873--1881},
  year={2025},
  publisher={Nature Publishing Group US New York}
}

@article{huang2025survey,
  title={A survey on hallucination in large language models: Principles, taxonomy, challenges, and open questions},
  author={Huang, Lei and Yu, Weijiang and Ma, Weitao and Zhong, Weihong and Feng, Zhangyin and Wang, Haotian and Chen, Qianglong and Peng, Weihua and Feng, Xiaocheng and Qin, Bing and others},
  journal={ACM Transactions on Information Systems},
  volume={43},
  number={2},
  pages={1--55},
  year={2025},
  publisher={ACM New York, NY}
}

@article{gallegos2024bias,
  title={Bias and fairness in large language models: A survey},
  author={Gallegos, Isabel O and Rossi, Ryan A and Barrow, Joe and Tanjim, Md Mehrab and Kim, Sungchul and Dernoncourt, Franck and Yu, Tong and Zhang, Ruiyi and Ahmed, Nesreen K},
  journal={Computational linguistics},
  volume={50},
  number={3},
  pages={1097--1179},
  year={2024},
  publisher={MIT Press 255 Main Street, 9th Floor, Cambridge, Massachusetts 02142, USA~...}
}

@article{chelli2024hallucination,
  title={Hallucination rates and reference accuracy of ChatGPT and bard for systematic reviews: comparative analysis},
  author={Chelli, Mika{\"e}l and Descamps, Jules and Lavou{\'e}, Vincent and Trojani, Christophe and Azar, Michel and Deckert, Marcel and Raynier, Jean-Luc and Clowez, Gilles and Boileau, Pascal and Ruetsch-Chelli, Caroline and others},
  journal={Journal of medical Internet research},
  volume={26},
  number={1},
  pages={e53164},
  year={2024},
  publisher={JMIR Publications Inc., Toronto, Canada}
}

@article{he2025survey,
  title={A survey of large language models for healthcare: from data, technology, and applications to accountability and ethics},
  author={He, Kai and Mao, Rui and Lin, Qika and Ruan, Yucheng and Lan, Xiang and Feng, Mengling and Cambria, Erik},
  journal={Information Fusion},
  volume={118},
  pages={102963},
  year={2025},
  publisher={Elsevier}
}

@article{zhao2026survey,
  title={A survey of large language models},
  author={Zhao, Wayne Xin and Zhou, Kun and Li, Junyi and Tang, Tianyi and Dong, Zican and Hou, Yupeng and Zhang, Beichen and Min, Yingqian and Zhang, Junjie and Liu, Peiyu and others},
  journal={Frontiers of Computer Science},
  volume={20},
  number={12},
  pages={2012627},
  year={2026},
  publisher={Springer}
}

@article{dahl2024large,
  title={Large legal fictions: Profiling legal hallucinations in large language models},
  author={Dahl, Matthew and Magesh, Varun and Suzgun, Mirac and Ho, Daniel E},
  journal={Journal of Legal Analysis},
  volume={16},
  number={1},
  pages={64--93},
  year={2024},
  publisher={Oxford University Press UK}
}

@article{hassija2024interpreting,
  title={Interpreting black-box models: a review on explainable artificial intelligence},
  author={Hassija, Vikas and Chamola, Vinay and Mahapatra, Atmesh and Singal, Abhinandan and Goel, Divyansh and Huang, Kaizhu and Scardapane, Simone and Spinelli, Indro and Mahmud, Mufti and Hussain, Amir},
  journal={Cognitive Computation},
  volume={16},
  number={1},
  pages={45--74},
  year={2024},
  publisher={Springer}
}

@article{das2025security,
  title={Security and privacy challenges of large language models: A survey},
  author={Das, Badhan Chandra and Amini, M Hadi and Wu, Yanzhao},
  journal={ACM Computing Surveys},
  volume={57},
  number={6},
  pages={1--39},
  year={2025},
  publisher={ACM New York, NY}
}

@article{nabben2024ai,
  title={AI as a constituted system: accountability lessons from an LLM experiment},
  author={Nabben, Kelsie},
  journal={Data \& policy},
  volume={6},
  pages={e57},
  year={2024},
  publisher={Cambridge University Press}
}

@article{elliott2025evolving,
  title={Evolving generative AI: entangling the accountability relationship},
  author={Elliott, Marc T J and P, Deepak and Maccarthaigh, Muiris},
  journal={Digital Government: Research and Practice},
  volume={6},
  number={1},
  pages={1--13},
  year={2025},
  publisher={ACM New York, NY}
}

@article{novelli2024accountability,
  title={Accountability in artificial intelligence: what it is and how it works},
  author={Novelli, Claudio and Taddeo, Mariarosaria and Floridi, Luciano},
  journal={Ai \& Society},
  volume={39},
  number={4},
  pages={1871--1882},
  year={2024},
  publisher={Springer}
}

@article{ferdaus2026towards,
  title={Towards trustworthy AI: a review of ethical and robust large language models},
  author={Ferdaus, Md Meftahul and Abdelguerfi, Mahdi and Loup, Elias and N. Niles, Kendall and Pathak, Ken and Sloan, Steven},
  journal={ACM Computing Surveys},
  volume={58},
  number={7},
  pages={1--43},
  year={2026},
  publisher={ACM New York, NY}
}

@article{zeiser2024owning,
  title={Owning decisions: AI decision-support and the attributability-gap},
  author={Zeiser, Jannik},
  journal={Science and Engineering Ethics},
  volume={30},
  number={4},
  pages={27},
  year={2024},
  publisher={Springer}
}

@article{green2022flaws,
  title={The flaws of policies requiring human oversight of government algorithms},
  author={Green, Ben},
  journal={Computer Law \& Security Review},
  volume={45},
  pages={105681},
  year={2022},
  publisher={Elsevier}
}

@inproceedings{sterz2024quest,
  title={On the quest for effectiveness in human oversight: Interdisciplinary perspectives},
  author={Sterz, Sarah and Baum, Kevin and Biewer, Sebastian and Hermanns, Holger and Lauber-R{\"o}nsberg, Anne and Meinel, Philip and Langer, Markus},
  booktitle={Proceedings of the 2024 ACM Conference on Fairness, Accountability, and Transparency},
  pages={2495--2507},
  year={2024}
}

@article{mokander2024auditing,
  title={Auditing large language models: a three-layered approach},
  author={M{\"o}kander, Jakob and Schuett, Jonas and Kirk, Hannah Rose and Floridi, Luciano},
  journal={AI and Ethics},
  volume={4},
  number={4},
  pages={1085--1115},
  year={2024},
  publisher={Springer}
}

@article{jiao2025generative,
  title={Generative AI and LLMs in industry: A text-mining analysis and critical evaluation of guidelines and policy statements across fourteen industrial sectors},
  author={Jiao, Junfeng and Afroogh, Saleh and Chen, Kevin and Atkinson, David and Dhurandhar, Amit},
  journal={arXiv preprint arXiv:2501.00957},
  year={2025}
}

@article{lognoul2025regulation,
  title={Regulation (EU) 2024/1689 laying down harmonised rules on artificial intelligence (Artificial Intelligence Act--AI Act)},
  author={Lognoul, Michael},
  journal={Revue du Droit des Technologies de l'information},
  volume={2025},
  number={3-4},
  pages={145--189},
  year={2025},
  publisher={Larcier}
}

@article{ai2023artificial,
  title={Artificial intelligence risk management framework (AI RMF 1.0)},
  author={AI, NIST},
  journal={\url{https://nvlpubs.nist.gov/nistpubs/ai/nist.ai}},
  pages={100--1},
  volume={1},
  number={1},
  year={2023}
}

@article{ai2024artificial,
  title={Artificial intelligence risk management framework: Generative artificial intelligence profile},
  author={AI, NIST},
  journal={NIST Trustworthy and Responsible AI Gaithersburg, MD, USA},
  year={2024}
}

@manual{iso2023iso,
  author = {{International Organization for Standardization} and {International Electrotechnical Commission}},
  title = {{ISO/IEC 42001:2023} Information technology Artificial intelligence Management system},
  organization = {ISO},
  address = {Geneva, Switzerland},
  year = {2023},
  note = {\url{https://www.iso.org/standard/81230.html}}
}

@inproceedings{hacker2023regulating,
  title={Regulating ChatGPT and other large generative AI models},
  author={Hacker, Philipp and Engel, Andreas and Mauer, Marco},
  booktitle={Proceedings of the 2023 ACM conference on fairness, accountability, and transparency},
  pages={1112--1123},
  year={2023}
}

@article{veale2021demystifying,
  title={Demystifying the draft EU artificial intelligence act},
  author={Veale, Michael and Borgesius, Frederik Zuiderveen},
  journal={arXiv preprint arXiv:2107.03721},
  year={2021}
}

@inproceedings{wieringa2020account,
  title={What to account for when accounting for algorithms: a systematic literature review on algorithmic accountability},
  author={Wieringa, Maranke},
  booktitle={Proceedings of the 2020 conference on fairness, accountability, and transparency},
  pages={1--18},
  year={2020}
}

@article{hollanek2023ai,
  title={AI transparency: a matter of reconciling design with critique},
  author={Hollanek, Tomasz},
  journal={Ai \& Society},
  volume={38},
  number={5},
  pages={2071--2079},
  year={2023},
  publisher={Springer}
}

@article{wang2025survey,
  title={Survey on factuality in large language models},
  author={Wang, Cunxiang and Liu, Xiaoze and Yue, Yuanhao and Guo, Qipeng and Hu, Xiangkun and Tang, Xiangru and Zhang, Tianhang and Jiayang, Cheng and Yao, Yunzhi and Hu, Xuming and others},
  journal={ACM Computing Surveys},
  volume={58},
  number={1},
  pages={1--37},
  year={2025},
  publisher={ACM New York, NY}
}

@article{weidinger2023sociotechnical,
  title={Sociotechnical safety evaluation of generative ai systems},
  author={Weidinger, Laura and Rauh, Maribeth and Marchal, Nahema and Manzini, Arianna and Hendricks, Lisa Anne and Mateos-Garcia, Juan and Bergman, Stevie and Kay, Jackie and Griffin, Conor and Bariach, Ben and others},
  journal={arXiv preprint arXiv:2310.11986},
  year={2023}
}

@article{bommasani2023foundation,
  title={The foundation model transparency index},
  author={Bommasani, Rishi and Klyman, Kevin and Longpre, Shayne and Kapoor, Sayash and Maslej, Nestor and Xiong, Betty and Zhang, Daniel and Liang, Percy},
  journal={arXiv preprint arXiv:2310.12941},
  volume={N/A},
  number={N/A},
  year={2023}
}

@article{roberts2024global,
  title={Global AI governance: barriers and pathways forward},
  author={Roberts, Huw and Hine, Emmie and Taddeo, Mariarosaria and Floridi, Luciano},
  journal={International Affairs},
  volume={100},
  number={3},
  pages={1275--1286},
  year={2024},
  publisher={Oxford University Press}
}

@article{ren2024reconciling,
  title={Reconciling the contrasting narratives on the environmental impact of large language models},
  author={Ren, Shaolei and Tomlinson, Bill and Black, Rebecca W and Torrance, Andrew W},
  journal={Scientific Reports},
  volume={14},
  number={1},
  pages={26310},
  year={2024},
  publisher={Nature Publishing Group UK London}
}

@article{liu2024green,
  title={Green AI: exploring carbon footprints, mitigation strategies, and trade offs in large language model training},
  author={Liu, Vivian and Yin, Yiqiao},
  journal={Discover Artificial Intelligence},
  volume={4},
  number={1},
  pages={49},
  year={2024},
  publisher={Springer}
}

@article{hagendorff2024mapping,
  title={Mapping the Ethics of Generative AI: A Comprehensive Scoping Review: T. Hagendorff},
  author={Hagendorff, Thilo},
  journal={Minds and Machines},
  volume={34},
  number={4},
  pages={39},
  year={2024},
  publisher={Springer}
}

@article{petrov2024dager,
  title={Dager: Exact gradient inversion for large language models},
  author={Petrov, Ivo and Dimitrov, Dimitar I and Baader, Maximilian and M{\"u}ller, Mark N and Vechev, Martin},
  journal={Advances in neural information processing systems},
  volume={37},
  pages={87801--87830},
  year={2024}
}

@inproceedings{hasanain2024large,
  title={Large language models for propaganda span annotation},
  author={Hasanain, Maram and Ahmad, Fatema and Alam, Firoj},
  booktitle={Findings of the Association for Computational Linguistics: EMNLP 2024},
  pages={14522--14532},
  year={2024}
}

@inproceedings{manakul2023selfcheckgpt,
  title={Selfcheckgpt: Zero-resource black-box hallucination detection for generative large language models},
  author={Manakul, Potsawee and Liusie, Adian and Gales, Mark},
  booktitle={Proceedings of the 2023 conference on empirical methods in natural language processing},
  pages={9004--9017},
  year={2023}
}

@article{kuhn2023semantic,
  title={Semantic uncertainty: Linguistic invariances for uncertainty estimation in natural language generation},
  author={Kuhn, Lorenz and Gal, Yarin and Farquhar, Sebastian},
  journal={arXiv preprint arXiv:2302.09664},
  year={2023}
}

@article{mishra2024fine,
  title={Fine-grained hallucination detection and editing for language models},
  author={Mishra, Abhika and Asai, Akari and Balachandran, Vidhisha and Wang, Yizhong and Neubig, Graham and Tsvetkov, Yulia and Hajishirzi, Hannaneh},
  journal={arXiv preprint arXiv:2401.06855},
  year={2024}
}

@inproceedings{niu2024ragtruth,
  title={Ragtruth: A hallucination corpus for developing trustworthy retrieval-augmented language models},
  author={Niu, Cheng and Wu, Yuanhao and Zhu, Juno and Xu, Siliang and Shum, KaShun and Zhong, Randy and Song, Juntong and Zhang, Tong},
  booktitle={Proceedings of the 62nd Annual Meeting of the Association for Computational Linguistics (Volume 1: Long Papers)},
  pages={10862--10878},
  year={2024}
}

@article{farquhar2024detecting,
  title={Detecting hallucinations in large language models using semantic entropy},
  author={Farquhar, Sebastian and Kossen, Jannik and Kuhn, Lorenz and Gal, Yarin},
  journal={Nature},
  volume={630},
  number={8017},
  pages={625--630},
  year={2024},
  publisher={Nature Publishing Group UK London}
}

@article{schuett2025three,
  title={Three lines of defense against risks from AI},
  author={Schuett, Jonas},
  journal={AI \& SOCIETY},
  volume={40},
  number={2},
  pages={493--507},
  year={2025},
  publisher={Springer}
}

@article{mokander2023operationalising,
  title={Operationalising AI governance through ethics-based auditing: an industry case study},
  author={M{\"o}kander, Jakob and Floridi, Luciano},
  journal={AI and Ethics},
  volume={3},
  number={2},
  pages={451--468},
  year={2023},
  publisher={Springer}
}

@article{dyba2005evidence,
  title={Evidence-based software engineering for practitioners},
  author={Dyba, Tore and Kitchenham, Barbara A and Jorgensen, Magne},
  journal={IEEE software},
  volume={22},
  number={1},
  pages={58--65},
  year={2005},
  publisher={IEEE}
}

@misc{sohrabi2021prisma,
  title={PRISMA 2020 statement: What's new and the importance of reporting guidelines},
  author={Sohrabi, Catrin and Franchi, Thomas and Mathew, Ginimol and Kerwan, Ahmed and Nicola, Maria and Griffin, Michelle and Agha, Maliha and Agha, Riaz},
  journal={International Journal of Surgery},
  volume={88},
  pages={105918},
  year={2021},
  publisher={Elsevier}
}

@article{page2021prisma,
  title={PRISMA 2020 explanation and elaboration: updated guidance and exemplars for reporting systematic reviews},
  author={Page, Matthew J and Moher, David and Bossuyt, Patrick M and Boutron, Isabelle and Hoffmann, Tammy C and Mulrow, Cynthia D and Shamseer, Larissa and Tetzlaff, Jennifer M and Akl, Elie A and Brennan, Sue E and others},
  journal={bmj},
  volume={372},
  year={2021},
  publisher={British Medical Journal Publishing Group}
}

@article{arvanitou2021software,
  title={Software engineering practices for scientific software development: A systematic mapping study},
  author={Arvanitou, Elvira-Maria and Ampatzoglou, Apostolos and Chatzigeorgiou, Alexander and Carver, Jeffrey C},
  journal={Journal of Systems and Software},
  volume={172},
  pages={110848},
  year={2021},
  publisher={Elsevier}
}

@article{xiao2019guidance,
  title={Guidance on conducting a systematic literature review},
  author={Xiao, Yu and Watson, Maria},
  journal={Journal of planning education and research},
  volume={39},
  number={1},
  pages={93--112},
  year={2019},
  publisher={SAGE Publications Sage CA: Los Angeles, CA}
}

@article{da2024survey,
  title={A survey of large language models in cybersecurity},
  author={da Silva, Gabriel de Jesus Coelho and Westphall, Carlos Becker},
  journal={arXiv preprint arXiv:2402.16968},
  year={2024}
}

@inproceedings{edenberg2023disambiguating,
  title={Disambiguating algorithmic bias: From neutrality to justice},
  author={Edenberg, Elizabeth and Wood, Alexandra},
  booktitle={Proceedings of the 2023 AAAI/ACM Conference on AI, Ethics, and Society},
  pages={691--704},
  year={2023}
}

@article{smuha2021race,
  title={From a `race to AI'to a `race to AI regulation': regulatory competition for artificial intelligence},
  author={Smuha, Nathalie A},
  journal={Law, Innovation and Technology},
  volume={13},
  number={1},
  pages={57--84},
  year={2021},
  publisher={Taylor \& Francis}
}

@article{vasconcelos2023explanations,
  title={Explanations can reduce overreliance on ai systems during decision-making},
  author={Vasconcelos, Helena and J{\"o}rke, Matthew and Grunde-McLaughlin, Madeleine and Gerstenberg, Tobias and Bernstein, Michael S and Krishna, Ranjay},
  journal={Proceedings of the ACM on Human-Computer Interaction},
  volume={7},
  number={CSCW1},
  pages={1--38},
  year={2023},
  publisher={ACM New York, NY, USA}
}

@article{inkpen2023advancing,
  title={Advancing human-AI complementarity: The impact of user expertise and algorithmic tuning on joint decision making},
  author={Inkpen, Kori and Chappidi, Shreya and Mallari, Keri and Nushi, Besmira and Ramesh, Divya and Michelucci, Pietro and Mandava, Vani and Vep{\v{r}}ek, Libu{\v{s}}e Hannah and Quinn, Gabrielle},
  journal={ACM Transactions on Computer-Human Interaction},
  volume={30},
  number={5},
  pages={1--29},
  year={2023},
  publisher={ACM New York, NY}
}

@article{yehudai2025survey,
  title={Survey on evaluation of llm-based agents},
  author={Yehudai, Asaf and Eden, Lilach and Li, Alan and Uziel, Guy and Zhao, Yilun and Bar-Haim, Roy and Cohan, Arman and Shmueli-Scheuer, Michal},
  journal={arXiv preprint arXiv:2503.16416},
  year={2025}
}

@article{leon2026lifecycle,
  title={Lifecycle-Based Governance to Build Reliable Ethical AI Systems},
  author={Leon, Maikel},
  journal={Systems Research and Behavioral Science},
  year={2026},
  publisher={Wiley Online Library}
}

@article{jongepier2022explanation,
  title={Explanation and Agency: exploring the normative-epistemic landscape of the ``Right to Explanation''},
  author={Jongepier, Fleur and Keymolen, Esther},
  journal={Ethics and Information Technology},
  volume={24},
  number={4},
  pages={49},
  year={2022},
  publisher={Springer}
}

@article{laux2024institutionalised,
  title={Institutionalised distrust and human oversight of artificial intelligence: towards a democratic design of AI governance under the European Union AI Act},
  author={Laux, Johann},
  journal={Ai \& Society},
  volume={39},
  number={6},
  pages={2853--2866},
  year={2024},
  publisher={Springer}
}

@article{abbas2025lending,
  title={Lending by algorithm: Fair or flawed? An information-theoretic view of credit decision pipelines},
  author={Abbas, Sayyed Khawar},
  journal={SN computer science},
  volume={6},
  number={6},
  pages={679},
  year={2025},
  publisher={Springer}
}

@inproceedings{raji2020closing,
  title={Closing the AI accountability gap: Defining an end-to-end framework for internal algorithmic auditing},
  author={Raji, Inioluwa Deborah and Smart, Andrew and White, Rebecca N and Mitchell, Margaret and Gebru, Timnit and Hutchinson, Ben and Smith-Loud, Jamila and Theron, Daniel and Barnes, Parker},
  booktitle={Proceedings of the 2020 conference on fairness, accountability, and transparency},
  pages={33--44},
  year={2020}
}

@article{gao2023retrieval,
  title={Retrieval-augmented generation for large language models: A survey},
  author={Gao, Yunfan and Xiong, Yun and Gao, Xinyu and Jia, Kangxiang and Pan, Jinliu and Bi, Yuxi and Dai, Yi and Sun, Jiawei and Wang, Meng and Wang, Haofen},
  journal={arXiv preprint arXiv:2312.10997},
  year={2023}
}

@article{inan2024llama,
  title={Llama guard: Llm-based input-output safeguard for human-ai conversations, 2023},
  author={Inan, Hakan and Upasani, Kartikeya and Chi, Jianfeng and Rungta, Rashi and Iyer, Krithika and Mao, Yuning and Tontchev, Michael and Hu, Qing and Fuller, Brian and Testuggine, Davide and others},
  journal={\url{https://arxiv.org/abs/2312.06674}},
  volume={2},
  number={6},
  pages={15},
  year={2024}
}

@inproceedings{rebedea2023nemo,
  title={Nemo guardrails: A toolkit for controllable and safe llm applications with programmable rails},
  author={Rebedea, Traian and Dinu, Razvan and Sreedhar, Makesh Narsimhan and Parisien, Christopher and Cohen, Jonathan},
  booktitle={Proceedings of the 2023 conference on empirical methods in natural language processing: system demonstrations},
  pages={431--445},
  address={Singapore},
  year={2023}
}

@article{jalan2026survey,
  title={Survey on llm safety: attacks, defenses, alignment, metrics, and guardrails},
  author={Jalan, Pratik and Abishethvarman, Vadivel and Chandna, Bhavik and Naseem, Usman},
  journal={Machine Learning},
  volume={115},
  number={6},
  pages={130},
  year={2026},
  publisher={Springer}
}

@article{lin2022teaching,
  title={Teaching models to express their uncertainty in words},
  author={Lin, Stephanie and Hilton, Jacob and Evans, Owain},
  journal={arXiv preprint arXiv:2205.14334},
  year={2022}
}

@inproceedings{kirchenbauer2023watermark,
  title={A watermark for large language models},
  author={Kirchenbauer, John and Geiping, Jonas and Wen, Yuxin and Katz, Jonathan and Miers, Ian and Goldstein, Tom},
  booktitle={International conference on machine learning},
  pages={17061--17084},
  year={2023},
  organization={PMLR}
}

@article{zhao2023provable,
  title={Provable robust watermarking for ai-generated text},
  author={Zhao, Xuandong and Ananth, Prabhanjan and Li, Lei and Wang, Yu-Xiang},
  journal={arXiv preprint arXiv:2306.17439},
  year={2023}
}

@article{asadollahi2026governing,
  title={Governing by design: algorithmic normativity, clinical standards, and health policy implications of AI in healthcare},
  author={Asadollahi, Ali},
  journal={AI and Ethics},
  volume={6},
  number={1},
  pages={119},
  year={2026},
  publisher={Springer}
}

@article{sharma2026human,
  title={Human-in-the-Loop Governance of Artificial Intelligence in Cardiology: From Ethical Principles to Operational Paradigms},
  author={Sharma, Kamal and Sharma, Praneel and Sharma, Pratyusha},
  journal={Indian Heart Journal},
  year={2026},
  publisher={Elsevier}
}

@article{singh2026architecting,
  title={Architecting Human-AI Systems for Effective Collaboration and Oversight: Making Sense of Human/AI-in/on/Over/Under/Along-the-Loop},
  author={Singh, Aditya and Szajnfarber, Zoe},
  journal={Systems Engineering},
  volume={29},
  number={2},
  pages={337--353},
  year={2026},
  publisher={Wiley Online Library}
}

@inproceedings{bansal2021does,
  title={Does the whole exceed its parts? the effect of ai explanations on complementary team performance},
  author={Bansal, Gagan and Wu, Tongshuang and Zhou, Joyce and Fok, Raymond and Nushi, Besmira and Kamar, Ece and Ribeiro, Marco Tulio and Weld, Daniel},
  booktitle={Proceedings of the 2021 CHI conference on human factors in computing systems},
  pages={1--16},
  year={2021}
}

@article{doshi2017towards,
  title={Towards a rigorous science of interpretable machine learning},
  author={Doshi-Velez, Finale and Kim, Been},
  journal={arXiv preprint arXiv:1702.08608},
  year={2017}
}

@article{van2026agentic,
  title={From agentic AI to AI-orchestrated organizations: Understanding the next surge in artificial intelligence},
  author={van Esch, Patrick},
  journal={Business Horizons},
  year={2026},
  publisher={Elsevier}
}

@article{hadley2025investigating,
  title={Investigating algorithm review boards for organizational responsible artificial intelligence governance},
  author={Hadley, Emily and Blatecky, Alan and Comfort, Megan},
  journal={AI and Ethics},
  volume={5},
  number={3},
  pages={2485--2495},
  year={2025},
  publisher={Springer}
}

@article{ganguli2022red,
  title={Red teaming language models to reduce harms: Methods, scaling behaviors, and lessons learned},
  author={Ganguli, Deep and Lovitt, Liane and Kernion, Jackson and Askell, Amanda and Bai, Yuntao and Kadavath, Saurav and Mann, Ben and Perez, Ethan and Schiefer, Nicholas and Ndousse, Kamal and others},
  journal={arXiv preprint arXiv:2209.07858},
  year={2022}
}

@article{mateo2026reference,
  title={Reference architecture for the design and implementation of AI systems in manufacturing in conformity to ISO/IEC 42001},
  author={Mateo-Casali, MA and Maa{\ss}en, Max and Heymann, Henrik and Boza, Andr{\'e}s and Fraile, Francisco and Leyendecker, Lars and Grunert, Dennis and Schmitt, Robert H},
  journal={International Journal of Computer Integrated Manufacturing},
  pages={1--23},
  year={2026},
  publisher={Taylor \& Francis}
}

@inproceedings{mitchell2019model,
  title={Model cards for model reporting},
  author={Mitchell, Margaret and Wu, Simone and Zaldivar, Andrew and Barnes, Parker and Vasserman, Lucy and Hutchinson, Ben and Spitzer, Elena and Raji, Inioluwa Deborah and Gebru, Timnit},
  booktitle={Proceedings of the conference on fairness, accountability, and transparency},
  pages={220--229},
  year={2019}
}

@article{gebru2021datasheets,
  title={Datasheets for datasets},
  author={Gebru, Timnit and Morgenstern, Jamie and Vecchione, Briana and Vaughan, Jennifer Wortman and Wallach, Hanna and Iii, Hal Daum{\'e} and Crawford, Kate},
  journal={Communications of the ACM},
  volume={64},
  number={12},
  pages={86--92},
  year={2021},
  publisher={ACM New York, NY, USA}
}

@inproceedings{dominguez2024mapping,
  title={Mapping the individual, social and biospheric impacts of Foundation Models},
  author={Dom{\'\i}nguez Hern{\'a}ndez, Andr{\'e}s and Krishna, Shyam and Perini, Antonella Maia and Katell, Michael and Bennett, SJ and Borda, Ann and Hashem, Youmna and Hadjiloizou, Semeli and Mahomed, Sabeehah and Jayadeva, Smera and others},
  booktitle={Proceedings of the 2024 ACM Conference on Fairness, Accountability, and Transparency},
  pages={776--796},
  year={2024}
}

@article{wang2026agent,
  title={From Agent Traces to Trust: Evidence Tracing and Execution Provenance in LLM Agents},
  author={Wang, Yiqi and Zhang, Jiaqi and Cai, Taotao and Liu, Zirui and Sun, Qingqiang and Sun, Zequn and Wu, Zhangkai and Zhang, Mingkai and Zhu, Yanming},
  journal={arXiv preprint arXiv:2606.04990},
  year={2026}
}

@article{naser2025auditing,
  title={Auditing the Shadows: A Review of Methods to Detect Shared Training Data in Large Language Models},
  author={Naser, MZ},
  journal={ACM Computing Surveys},
  volume={58},
  number={7},
  pages={1--34},
  year={2025},
  publisher={ACM New York, NY}
}

@article{saini2026regulatory,
  title={Regulatory challenges and opportunities: A review of US food and drug Administration-Approved artificial intelligence and machine Learning-Enabled cardiovascular devices},
  author={Saini, Mahima and Kc, Grishma and Williams, Adrian J and Coplan, Paul M and Gressler, Laura E},
  journal={Therapeutic Innovation \& Regulatory Science},
  volume={60},
  number={2},
  pages={393--422},
  year={2026},
  publisher={Springer}
}

@article{nollen2026artificial,
  title={Artificial Intelligence for Regulatory Evidence: A Systematic Document Analysis of European Medicines Agency Regulatory Advice and Public Reports},
  author={Nollen, Lucina-May and van Westen, Gerard JP and Westman, Gabriel and Pasmooij, Anna MG},
  journal={Clinical Pharmacology \& Therapeutics},
  year={2026},
  publisher={Wiley Online Library}
}

@book{united2023artificial,
  title={Artificial intelligence and the future of teaching and learning: Insights and recommendations},
  author={United States. Office of Educational Technology},
  year={2023},
  publisher={US Department of Education, Office of Educational Technology}
}

@book{holmes2023guidance,
  title={Guidance for generative AI in education and research},
  author={Holmes, Wayne and Miao, Fengchun and others},
  year={2023},
  publisher={Unesco Publishing}
}

@article{ali2026operationalising,
  title={Operationalising Information Security Management: A Procedural Framework Analysis of ISO/IEC 27001: 2022 Implementation in a Financial-Technology Organisation},
  author={Ali, Ratul},
  journal={arXiv preprint arXiv:2604.23230},
  year={2026}
}

@inproceedings{greshake2023not,
  title={Not what you've signed up for: Compromising real-world llm-integrated applications with indirect prompt injection},
  author={Greshake, Kai and Abdelnabi, Sahar and Mishra, Shailesh and Endres, Christoph and Holz, Thorsten and Fritz, Mario},
  booktitle={Proceedings of the 16th ACM workshop on artificial intelligence and security},
  pages={79--90},
  year={2023}
}

@article{yang2024assessing,
  title={Assessing adversarial robustness of large language models: An empirical study},
  author={Yang, Zeyu and Meng, Zhao and Zheng, Xiaochen and Wattenhofer, Roger},
  journal={arXiv preprint arXiv:2405.02764},
  year={2024}
}

@article{polemi2023multilayer,
  title={Multilayer framework for good cybersecurity practices for AI},
  author={Polemi, Nineta and Pra{\c{c}}a, Isabel},
  journal={ENISA, Tech. Rep.},
  year={2023}
}

@article{liang2024empowering,
  title={Empowering large language models to set up a knowledge retrieval indexer via self-learning},
  author={Liang, Xun and Niu, Simin and Zhang, Sensen and Song, Shichao and Wang, Hanyu and Yang, Jiawei and Xiong, Feiyu and Tang, Bo and Xi, Chenyang and others},
  journal={arXiv preprint arXiv:2405.16933},
  year={2024}
}

@article{izacard2023atlas,
  title={Atlas: Few-shot learning with retrieval augmented language models},
  author={Izacard, Gautier and Lewis, Patrick and Lomeli, Maria and Hosseini, Lucas and Petroni, Fabio and Schick, Timo and Dwivedi-Yu, Jane and Joulin, Armand and Riedel, Sebastian and Grave, Edouard},
  journal={Journal of Machine Learning Research},
  volume={24},
  number={251},
  pages={1--43},
  year={2023}
}

@article{floridi2022capai,
  title={CapAI-A procedure for conducting conformity assessment of AI systems in line with the EU artificial intelligence act},
  author={Floridi, Luciano and Holweg, Matthias and Taddeo, Mariarosaria and Amaya, Javier and M{\"o}kander, Jakob and Wen, Yuni},
  journal={Available at SSRN 4064091},
  year={2022}
}

@article{sachan2024blockchain,
  title={Blockchain-based auditing of legal decisions supported by explainable AI and generative AI tools},
  author={Sachan, Swati and Liu, Xi},
  journal={Engineering Applications of Artificial Intelligence},
  volume={129},
  pages={107666},
  year={2024},
  publisher={Elsevier}
}

@inproceedings{li2023halueval,
  title={Halueval: A large-scale hallucination evaluation benchmark for large language models},
  author={Li, Junyi and Cheng, Xiaoxue and Zhao, Xin and Nie, Jian-Yun and Wen, Ji-Rong},
  booktitle={The 2023 Conference on Empirical Methods in Natural Language Processing},
  year={2023}
}

@inproceedings{lin2022truthfulqa,
  title={Truthfulqa: Measuring how models mimic human falsehoods},
  author={Lin, Stephanie and Hilton, Jacob and Evans, Owain},
  booktitle={Proceedings of the 60th annual meeting of the association for computational linguistics (volume 1: long papers)},
  pages={3214--3252},
  year={2022}
}

@article{min2023factscore,
  title={Factscore: Fine-grained atomic evaluation of factual precision in long form text generation},
  author={Min, Sewon and Krishna, Kalpesh and Lyu, Xinxi and Lewis, Mike and Yih, Wen-tau and Koh, Pang Wei and Iyyer, Mohit and Zettlemoyer, Luke and Hajishirzi, Hannaneh},
  journal={arXiv preprint arXiv:2305.14251},
  year={2023}
}

@inproceedings{dhamala2021bold,
  title={Bold: Dataset and metrics for measuring biases in open-ended language generation},
  author={Dhamala, Jwala and Sun, Tony and Kumar, Varun and Krishna, Satyapriya and Pruksachatkun, Yada and Chang, Kai-Wei and Gupta, Rahul},
  booktitle={Proceedings of the 2021 ACM conference on fairness, accountability, and transparency},
  pages={862--872},
  year={2021}
}

@inproceedings{parrish2022bbq,
  title={BBQ: A hand-built bias benchmark for question answering},
  author={Parrish, Alicia and Chen, Angelica and Nangia, Nikita and Padmakumar, Vishakh and Phang, Jason and Thompson, Jana and Htut, Phu Mon and Bowman, Samuel R},
  booktitle={Findings of the Association for Computational Linguistics: ACL 2022},
  pages={2086--2105},
  year={2022}
}

@inproceedings{smith2022m,
  title={``I'm sorry to hear that'': Finding new biases in language models with a holistic descriptor dataset},
  author={Smith, Eric Michael and Hall, Melissa and Kambadur, Melanie and Presani, Eleonora and Williams, Adina},
  booktitle={Proceedings of the 2022 conference on empirical methods in natural language processing},
  pages={9180--9211},
  year={2022}
}

@article{shahin2026benchmarking,
  title={Benchmarking LLAMA Model Security Against OWASP Top 10 For LLM Applications},
  author={Shahin, Nourin and Alsmadi, Izzat},
  journal={arXiv preprint arXiv:2601.19970},
  year={2026}
}

@article{bovens2007analysing,
  title={Analysing and assessing accountability: A conceptual framework 1},
  author={Bovens, Mark},
  journal={European law journal},
  volume={13},
  number={4},
  pages={447--468},
  year={2007},
  publisher={Wiley Online Library}
}

@article{braun2006using,
  title={Using thematic analysis in psychology},
  author={Braun, Virginia and Clarke, Victoria},
  journal={Qualitative research in psychology},
  volume={3},
  number={2},
  pages={77--101},
  year={2006},
  publisher={Taylor \& Francis}
}

@misc{european2020guidelines,
  title={Guidelines on loan origination and monitoring},
  author={European Banking Authority},
  year={2020},
  publisher={European Banking Authority Paris}
}

@article{kitchenham2009systematic,
  title={Systematic literature reviews in software engineering--a systematic literature review},
  author={Kitchenham, Barbara and Brereton, O Pearl and Budgen, David and Turner, Mark and Bailey, John and Linkman, Stephen},
  journal={Information and software technology},
  volume={51},
  number={1},
  pages={7--15},
  year={2009},
  publisher={Elsevier}
}

@inproceedings{weidinger2022taxonomy,
  title={Taxonomy of risks posed by language models},
  author={Weidinger, Laura and Uesato, Jonathan and Rauh, Maribeth and Griffin, Conor and Huang, Po-Sen and Mellor, John and Glaese, Amelia and Cheng, Myra and Balle, Borja and Kasirzadeh, Atoosa and others},
  booktitle={Proceedings of the 2022 ACM conference on fairness, accountability, and transparency},
  pages={214--229},
  year={2022}
}

@article{vaswani2017attention,
  title={Attention is all you need},
  author={Vaswani, Ashish and Shazeer, Noam and Parmar, Niki and Uszkoreit, Jakob and Jones, Llion and Gomez, Aidan N and Kaiser, {\L}ukasz and Polosukhin, Illia},
  journal={Advances in neural information processing systems},
  volume={30},
  year={2017}
}

@book{united2024advancing,
  title={Advancing governance, innovation, and risk management for agency use of artificial intelligence},
  author={United States. Office of Management and Budget},
  year={2024},
  publisher={Office of Management and Budget}
}

@article{agrawal2026ai,
  title={AI Regulation in US States: Lessons Learned and Key Takeaways},
  author={Agrawal, Lavlin and Mulgund, Pavankumar and DaSouza, Richelle Oakley and Bhaya, Kavita and Singh, Raghvendra},
  journal={Communications of the ACM},
  volume={69},
  number={6},
  pages={68--77},
  year={2026},
  publisher={ACM New York, NY, USA}
}

@article{bureau2022consumer,
  title={Consumer Financial Protection Circular 2022-03: Adverse action notification requirements in connection with credit decisions based on complex algorithms},
  author={Bureau, Consumer Financial Protection},
  journal={June. Available online: https://www. govinfo. gov/content/pkg/FR-2022-06-14/pdf/2022-12729. pdf (accessed on 16 May 2025)},
  year={2022}
}

@article{morandin2023recommendation,
  title={Recommendation of the OECD council on artificial intelligence: inequality and inclusion1},
  author={Morand{\'\i}n-Ahuerma, Fabio},
  journal={Principios normativos para una {\'e}tica de la inteligencia artificial},
  pages={95--102},
  year={2023}
}

@incollection{huang2024genai,
  title={Genai application level security},
  author={Huang, Ken and Huang, Grace and Dawson, Adam and Wu, Daniel},
  booktitle={Generative AI security: Theories and practices},
  pages={199--237},
  year={2024},
  publisher={Springer}
}

@article{smuha2025regulation,
  title={Regulation 2024/1689 of the Eur. Parl. \& Council of June 13, 2024 (eu artificial intelligence act)},
  author={Smuha, Nathalie A},
  journal={International Legal Materials},
  volume={64},
  number={5},
  pages={1234--1381},
  year={2025},
  publisher={Cambridge University Press}
}

@inproceedings{kalai2024calibrated,
  title={Calibrated language models must hallucinate},
  author={Kalai, Adam Tauman and Vempala, Santosh S},
  booktitle={Proceedings of the 56th Annual ACM Symposium on Theory of Computing},
  pages={160--171},
  year={2024}
}

@article{xu2024hallucination,
  title={Hallucination is inevitable: An innate limitation of large language models},
  author={Xu, Ziwei and Jain, Sanjay and Kankanhalli, Mohan},
  journal={arXiv preprint arXiv:2401.11817},
  year={2024}
}

\end{document}